\documentclass[10pt,journal,compsoc]{IEEEtran}
\usepackage{cite}
\usepackage{graphicx}
\usepackage{amsmath}
\usepackage{amssymb}
\usepackage{amsthm}
\usepackage{subcaption}
\usepackage{bbding,pifont}
\usepackage{algorithm}
\usepackage{algorithmic}
\usepackage{booktabs}
\usepackage{multirow}
\usepackage{enumitem}
\usepackage{mathtools}
\usepackage{mathrsfs}
\usepackage{mdframed}
\usepackage{longtable}
\usepackage[table,xcdraw]{xcolor}
\usepackage{soul}
\usepackage{hyperref}
\hypersetup{
    colorlinks=true,
    urlcolor=black,
    linkcolor=black,
    citecolor=black,
}

\newcommand{\ECS}{\mathrm{ECS}} 

\newcommand{\Mb}{\mathbf{M}}

\newcommand{\Xb}{\mathbf{X}}
\newcommand{\xb}{\mathbf{x}}

\theoremstyle{plain}
\newtheorem{theorem}{Theorem}[section]

\newtheorem{lemma}[theorem]{Lemma}

\theoremstyle{definition}
\newtheorem{definition}[theorem]{Definition}

\theoremstyle{remark}
\newtheorem{remark}[theorem]{Remark}

\def \polylog {\mathrm{polylog}}

\def \polylog {\mathrm{polylog}}

\def \polylog {\mathrm{polylog}}

\def \polylog {\mathrm{polylog}}

\begin{document}

\title{Shaping Initial State Prevents Modality Competition in Multi-modal Fusion: A Two-stage Scheduling Framework via Fast Partial Information Decomposition}
\author{Jiaqi Tang, Yinsong Xu, Yang Liu, Qingchao Chen
\IEEEcompsocitemizethanks{
\IEEEcompsocthanksitem Jiaqi Tang, Yinsong Xu, and Qingchao Chen are with the Institute of Medical Technology, Peking University Health Science Center, the National Institute of Health Data Science, and the State Key Laboratory of General Artificial Intelligence, Peking University, Beijing, 100191, China. E-mail: jiaqi\_tang@hsc.pku.edu.cn, xuyinsong@bupt.edu.cn, qingchao.chen@pku.edu.cn.
\IEEEcompsocthanksitem Yang Liu is with Wangxuan Institute of Computer Technology, Peking University, Beijing, 100080, China, E-mail: yangliu@pku.edu.cn.
}
}

\markboth{Journal of \LaTeX\ Class Files,~Vol.~14, No.~8, August~2015}%
{Shell \MakeLowercase{\textit{et al.}}: Bare Demo of IEEEtran.cls for Computer Society Journals}

\IEEEtitleabstractindextext{
\begin{abstract}
Multi-modal fusion often suffers from modality competition during joint training, where one modality dominates the learning process, leaving others under-optimized.
Overlooking the critical impact of the model's initial state, most existing methods address this issue during the joint learning stage.
In this study, we introduce a two-stage training framework to shape the initial states through unimodal training \textit{before the joint training.} 
First, we propose the concept of Effective Competitive Strength (ECS) to quantify a modality's competitive strength. Our theoretical analysis further reveals that properly shaping the initial ECS by unimodal training achieves a provably tighter error bound $O(1/K^2)$ over the $O(1/K)$ when competition happens, $K$ represents the class number.
However, ECS is computationally intractable in deep neural networks. 
To bridge this gap, we develop a framework comprising two core components: a fine-grained computable diagnostic metric and an asynchronous training controller. 
For the metric, we first prove that mutual information(MI) is a principled proxy for ECS. Considering MI is induced by per-modality marginals and thus treats each modality in isolation, we further propose \textbf{FastPID}, a computationally efficient and differentiable solver for partial information decomposition, which decomposes the joint distribution's information into fine-grained measurements: modality-specific uniqueness, redundancy, and synergy. 
Guided by these measurements, our asynchronous controller dynamically \textit{balances} modalities by monitoring uniqueness and \textit{locates} the ideal initial state to start joint training by tracking peak synergy.
Experiments on diverse benchmarks demonstrate that our method achieves state-of-the-art performance(an average gain of 7.70\% across four datasets). 
Our work establishes that shaping the pre-fusion models' initial state is a powerful strategy that eases competition before it starts, reliably unlocking synergistic multi-modal fusion. Code is available at \href{https://github.com/SPIresearch/FastPID_Scheduling}{https://github.com/SPIresearch/FastPID\_Scheduling}.

\end{abstract}

}

\maketitle

\IEEEpeerreviewmaketitle

\section{Introduction}


\begin{figure}[tp]
    \centering
    \includegraphics[width=\linewidth]{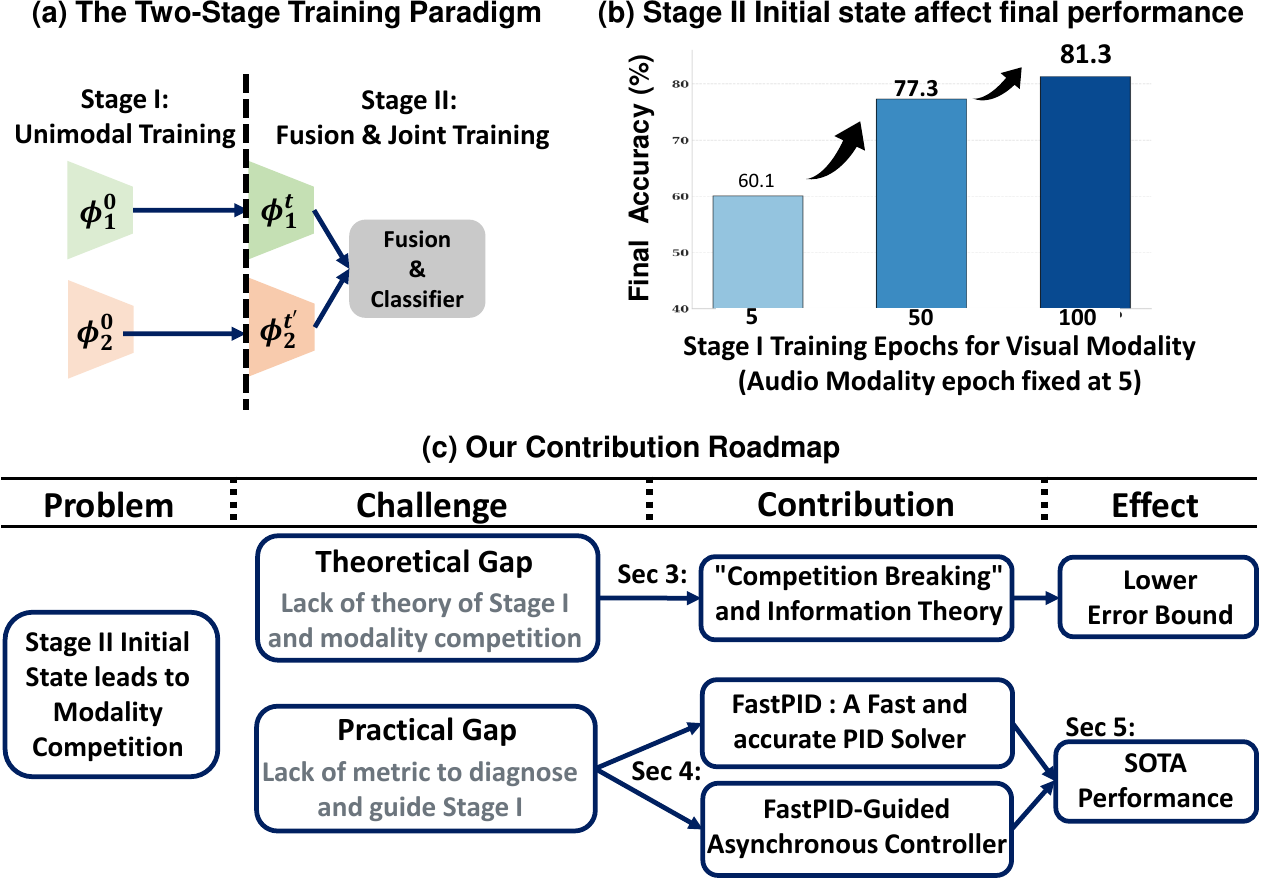}
    \caption{Illustration of our work's motivation and roadmap. (a) The two-stage training paradigm. (b) A prior experiment showed that the final performance is highly sensitive to the Stage II initial state. (c) Roadmap of our research, our method achieves a lower error bound and achieves the state-of-the-art performance.}
    \label{fig:teaser}
\end{figure}

Inspired by human cognition, where understanding comes through multiple sensory inputs~\cite{gazzaniga2006cognitive,jiao2024comprehensive}, multi-modal learning(MML) has become fundamental in various applications ranging from autonomous driving to embodied AI~\cite{ramachandram2017deep,huang2022multi,tang2023comparative}. Despite numerous advances, current MML approaches mainly rely on joint training paradigms, where features from unimodal encoders are fused into a shared representation and optimized by a uniform objective~\cite{baltruvsaitis2018multimodal}. However, this approach often leads to \textit{modality competition}~\cite{huang2022modality}, a detrimental training problem where certain modalities dominate the learning process while leaving others under-optimized, leading to limited overall performances~\cite{wei2024fly,yang2025learning}.

To formally analyze and address this challenge, we conceptualize the MML process through a two-stage framework, as illustrated in Figure~\ref{fig:teaser}(a): Stage I, a unimodal training stage that shapes the initial state of the later Stage, and Stage II, the subsequent fusion and joint training stage.
Within this framework, the vast majority of existing methods adopt the \textit{intervene-during-fusion paradigm} to address modality competition, where they focus only on Stage II, attempting to ease the competition \textit{during or after} it has emerged. 
These strategies can be broadly categorized as: (1)~\textit{optimization-level strategies}, which directly modulate gradients from each modality during parameter updates~\cite{peng2022balanced,li2023boosting,fanPMRPrototypicalModal2023}; (2)~\textit{objectives-level strategies}, such as modifying the multimodal loss function~\cite{xu2023mmcosine}, incorporating unimodal loss~\cite{du2023uni, weiMMParetoBoostingMultimodal2024}. And (3)~\textit{data-level strategies}, which employ sophisticated re-weighting or re-sampling schemes to amplify the influence of samples where a modality's contribution is weakest~\cite{weiEnhancingMultimodalCooperation2024}. 
However, the above works solely focus on stage II and overlook a critical factor: \textit{the model's initial state of Stage II, a condition inherited from Stage I training schedule.} This oversight is significant, as our experiments reveal that \textbf{the final performance is highly sensitive to the Stage II initial state configured by the Stage I schedule.} For instance, on the CREMA-D~\cite{cao2014crema} audio-visual dataset, as shown in Figure~\ref{fig:teaser}(b), simply extending Stage I training of the visual modality from 1 to 100 epochs while keeping the audio Stage I fixed at 5 epochs, improves the final accuracy from 60.1\% to 81.3\%. A detailed analysis of this phenomenon is presented in Section~\ref{sec:coeffient}.
Although a few methods~\cite{wang2020makes, du2023uni,hua2024reconboost} incorporated Stage I unimodal training, they generally treat it as a straightforward warm-up module, lacking both a theoretical understanding of how the Stage I schedule impacts the final performance and a principled mechanism to optimize it accordingly.

Consequently, a fundamental bottleneck problem persists: \textbf{there is no theoretical analysis about how Stage II's initial states, configured by the Stage I scheduling, \textit{affect} modality competition, nor a principled framework to control the Stage I training schedule to mitigate subsequent modality competition.} In this paper, we confront this challenge by establishing a coherent theoretical and computational implementation framework that transforms the Stage I training from a straightforward warm-up module into a schedule-guided, competition-aware procedure.

First, to establish the missing theoretical foundation for analyzing how the Stage I schedule affects subsequent modality competition, we build our theoretical analysis based on a simplified data model(see Section~\ref{sec:formulation}). 
We first propose the concept of Effective Competitive Strength (ECS) to quantify its competitive strength among modalities during the training. Our further analysis demonstrates that the initial ECS values of the Stage II initial state act as the critical trigger for the ``winner-takes-all'' modality domination~\cite{an2017analyzing, huang2022modality}. This insight allows us to precisely delineate four predictable outcomes based on the Stage~I training schedule: three failure modes, namely \textit{Amplified Competition}, \textit{Persistent Competition}, and \textit{Reversed Competition}, and a desirable state \textbf{\textit{Competition Breaking}}, which is elusive and demands an accurately scheduled Stage I training to establish a balanced initial ECS. This successful initialization leads to a provably and quadratically lower test error bound of $\mathcal{O}(1/K^2)$, a significant improvement over the $\Omega(1/K)$ bound observed when competition occurs.

Second, as the ECS is defined within a simplified model, it is intractable to compute for real-world deep neural networks, leading to a gap between theory and computational implementation. That raises a critical question: \textit{How can we locate the "Competition Breaking" initial state for Stage II without resorting to a computationally prohibitive brute-force search?} 
To address this, we propose a two-stage scheduling framework to automate the Stage I training. The framework is composed of two core components: (1) a fine-grained, computable metric to diagnose modality interactions, and (2) an automated controller that steers the training schedule based on this diagnosis.

\textit{For the metric}, beyond the intractable ECS, we require a computable diagnostic measurement to quantify the dynamic interactions between modalities during Stage I. 
We first theoretically proves that mutual information(MI) could act as a proxy of ECS(Section~\ref{sec:formulation}). However, a critical limitation emerges: both ECS and its direct proxy, MI, are calculated from the coarse-grained \textit{marginal distributions} $p(X_r, y)$, analyzing each modality in isolation.
To overcome this, we propose FastPID, a computationally efficient and differentiable solver for the Partial Information Decomposition (PID) framework. The PID dissects the cross-modal information of the \textit{joint distribution} $p(X_1, X_2, Y)$ into fine-grained components: modality uniqueness (exposing dominance), shared redundancy, and emergent synergy (quantifying collaboration), as illustrated in Section~\ref{sec:pid_bg}.
While conventional PID solvers are too slow and non-differentiable, our proposed FastPID overcomes these limitations through a two-phase hybrid approach that combines a rapid, closed-form analytical proxy with a stable, optimization-based refinement, enabling the fast and stable estimation of modality interactions.

\textit{For the control mechanism,} we propose an asynchronous controller that periodically uses this diagnosis to control the unimodal training schedule. Specifically, it uses the modality uniqueness ratio terms to dynamically alternate training between the encoders, selectively pausing updates for the dominant one to allow the other to catch up. Meanwhile, it tracks the growth of synergistic information, using its peak as a signal to determine the moment to conclude Stage I as the Stage II initial state.

Different from the prevailing ``intervene-during-fusion" paradigm that reacts to competition during Stage~II, we proactively shape its initial state by scheduling the Stage~I training. It is facilitated by our FastPID diagnostic module, which provides critical, forward-looking insight even before joint training commences. By allowing us to predict the dynamics of the subsequent fusion phase, the controller could select an ideal fusion point rather than resorting to a costly, brute-force search. Ultimately, this transforms Stage~I training from an unguided, straightforward warm-up module into a principled, predictive, and foundational phase of multi-modal fusion, shaping initial states for synergistic collaboration rather than competitive conflict.

Our contributions are summarized as follows:
\begin{itemize}
    \item  We establish a rigorous theoretical framework that formalizes the critical impact of Stage~I uni-modal training on modality competition. Our analysis delineates four distinct outcomes of Stage~I training, revealing an ideal ``Competition Breaking'' state for achieving balanced and synergistic multi-modal fusion.

    \item We propose \textbf{FastPID}, a computationally efficient and differentiable Partial Information Decomposition (PID) solver.
    It overcomes the bottlenecks of conventional methods, enabling a real-time diagnostic tool to guide the training of MML deep neural networks.

    \item We propose a two-stage scheduling framework that leverages an asynchronous controller to i) dynamically balance uni-modal training by monitoring modality-specific \textit{uniqueness}, and ii) identify the ideal moment for joint fusion by tracking emergent cross-modal \textit{synergy}.

    \item  Comprehensive experiments on diverse multi-modal benchmarks demonstrate that our method significantly outperforms existing approaches and establishes a new state-of-the-art (an average gain of 7.70\% across four datasets), validating the power of shaping the pre-fusion landscape. 
\end{itemize}



\section{Related Work and Background}

\subsection{Imbalanced Multimodal Learning}

The challenge of modality competition has been primarily addressed through strategies that dynamically balance learning efforts during the joint training phase. These approaches intervene at various levels. \textbf{Optimization-level strategies} directly modulate gradients to enhance under-optimized modalities~\cite{peng2022balanced, li2023boosting} or leverage class prototypes to rebalance learning by accelerating slower modalities~\cite{fanPMRPrototypicalModal2023}. \textbf{Objectives-level strategies} refine the loss functions, for instance by incorporating auxiliary uni-modal losses to ensure individual encoders are adequately trained~\cite{du2023uni}, designing novel objectives like a multi-modal cosine loss~\cite{xu2023mmcosine}, or finding Pareto-optimal gradient directions that benefit all modalities without conflict~\cite{weiMMParetoBoostingMultimodal2024}. Finally, \textbf{Data-level strategies} employ sophisticated re-weighting or re-sampling schemes, often guided by sample-level modality valuation, to amplify the influence of under-contributing modalities and ensure balanced contributions during training~\cite{weiEnhancingMultimodalCooperation2024}.

While effective, these approaches share a fundamental characteristic: they are designed to address the modality competition during the joint training stage, attempting to correct an existing competition. Our work diverges fundamentally from this paradigm by proposing a two-stage framework that prevents the competition from occurring by using the unimodal training to shape the model's initial state before the joint training.

\subsection{Pretraining in Multimodal Competition}
In contrast to the extensive work on joint training, the preparatory unimodal training phase (Stage 1) is far less understood, with its impact often considered trivial or even detrimental. Early investigations into modality imbalance found that a standard uni-modal pre-training and fine-tuning paradigm failed to offer improvements over end-to-end training~\cite{wang2020makes}, highlighting the non-trivial nature of the problem.
Further analysis suggested that this is because joint training is necessary to learn synergistic \textit{paired features} that cannot be captured by uni-modal encoders in isolation~\cite{du2023uni}. While some methods incorporate unimodal training elements, such as the alternating-boosting optimization in ReconBoost~\cite{hua2024reconboost}, they only train each modality until it achieves the best accuracy, and use this as a warm-up. Their focus remains on the dynamics within the joint training loop rather than systematically preparing the model's initial state. 

Consequently, a critical gap exists: there is no principled framework for guiding Stage I unimodal training to create an ideal initial state for Stage 2 fusion. The relationship between the schedule of unimodal training and the emergence of competition has not been formally characterized, nor are there methods to identify the ideal transition point. Our work directly addresses this gap by providing both a theoretical analysis of the Stage 1 impact and a FastPID-guided scheduling methodology to guide this crucial preparatory phase. 

\subsection{Background: Partial Information Decomposition}
\label{sec:pid_bg}

\begin{figure}[tp]
    \centering
    \includegraphics[width=0.95\linewidth]{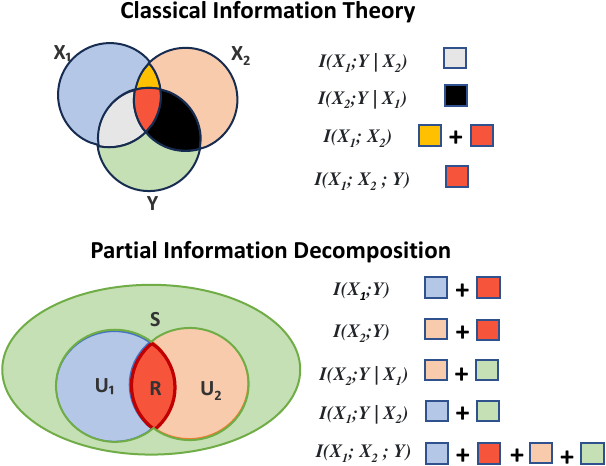}
    \caption{A comparison between Classical Information Theory and PID. PID (bottom) provides a principled decomposition of information from two sources ($X_1, X_2$) about a target ($Y$) into four non-negative atoms: unique information ($U_1, U_2$), shared redundancy ($R$), and emergent synergy ($S$).}
    \label{fig:PID}
\end{figure}

While classical information theory usually quantifies the information between two variables, its extension to three variables remains an open problem. As previous work~\cite{garner1962uncertainty,griffith2014quantifying,mcgill1954multivariate} defined $I(X_1; X_2; Y) = I(X_1; X_2) - I(X_1; X_2|Y)$, it can be both positive and negative, which makes it difficult to interpret as a measure of shared information, making it an unreliable tool for analysis.

\textbf{Partial Information Decomposition (PID)}
\label{sec:pidbg}
~\cite{williams2010nonnegative} is a framework designed to resolve this ambiguity by dissecting the total information $I(X_1, X_2; Y)$ into distinct, non-negative components~\cite{bertschinger2014quantifying}. Specifically, for two source variables $X_1$ and $X_2$ and a target $Y$, PID decomposes the information into four fundamental terms, illustrated in Figure~\ref{fig:PID}:
\begin{itemize}
    \item \textbf{Redundancy ($R$):} The information about $Y$ that is redundantly shared by both $X_1$ and $X_2$.
    \item \textbf{Unique Information ($U_1, U_2$):} The information about $Y$ that is exclusively present in $X_1$ ($U_1$) or in $X_2$ ($U_2$).
    \item \textbf{Synergy ($S$):} The new information about $Y$ that emerges only when $X_1$ and $X_2$ are considered together.
\end{itemize}
The formal definition of these components is rooted in an optimization over a set of distributions $\Delta_p$, which contains all distributions $q$ that preserve the marginal relationships from the true data distribution $p$:
\begin{equation}
    \Delta_p = \{q \mid q(x_i, y) = p(x_i, y) \text{ for } i \in \{1, 2\}\}
    \label{eq:marginal_constrait}
\end{equation}
where the PID terms are defined as:
\begin{align}
    R &= \max_{q \in \Delta_p} I_q(X_1; X_2; Y) \label{eq:pid_R} \\
    U_1 &= \min_{q \in \Delta_p} I_q(X_1; Y | X_2) \label{eq:pid_U1} \\
    U_2 &= \min_{q \in \Delta_p} I_q(X_2; Y | X_1) \label{eq:pid_U2} \\
    S &= I_p(X_1, X_2; Y) - \min_{q \in \Delta_p} I_q(X_1, X_2; Y) \label{eq:pid_S}
\end{align}

\begin{table}[htbp]
\centering
\caption{Summary of key notations used in the theoretical framework.}
\label{tab:notation}
\resizebox{\linewidth}{!}{%
\begin{tabular}{@{}ll@{}}
\toprule
\textbf{Symbol} & \textbf{Definition / Description} \\ \midrule
\multicolumn{2}{l}{\textit{\textbf{General Parameters}}} \\
$r \in \{1, 2\}$ & Index for a modality. \\
$K$ & Total number of classes. \\
$n$ & Number of samples in the dataset. \\
$d_r$ & Input feature dimension for modality $r$. \\
$y \in [K]$ & The ground-truth label for a sample. \\
\midrule
\multicolumn{2}{l}{\textit{\textbf{Network and Training}}} \\
$\varphi_{\mathcal{M}r}$ & Encoder for modality $r$. \\
$\mathcal{C}$ & Shared classifier module. \\
$w_{j,l,r}^{(t)}$ & Weight vector of the $l$-th neuron for class $j$ in modality $r$'s encoder at time $t$. \\
$\mathcal{L}_r, \mathcal{L}_{\text{joint}}$ & Uni-modal and multi-modal cross-entropy loss functions. \\
$T_{\text{pre}}^{(r)}, T_{\text{joint}}$ & Number of iterations for uni-modal pre-training and joint training. \\
$\sigma(\cdot)$ & Smoothed ReLU activation function. \\
$\beta, q$ & Parameters of the smoothed ReLU (threshold and exponent). \\
$\eta, \sigma_0$ & Learning rate and magnitude of weight initialization. \\
\midrule
\multicolumn{2}{l}{\textit{\textbf{Sparse Coding Data Model}}} \\
$\Xb^r$ & Input data for modality $r$. \\
$\Mb^r \in \mathbb{R}^{d_r \times K}$ & Unitary feature dictionary for modality $r$. \\
$z^r \in \mathbb{R}^K$ & Sparse vector. \\
$\xi^r$ & Isotropic noise vector. \\
\midrule
\multicolumn{2}{l}{\textit{\textbf{Core Theoretical Concepts}}} \\
$\Lambda_{j,r}^{(t)}$ & \textbf{Effective Competitive Strength (ECS)}: A modality's competitive power for class $j$. \\
$\Gamma_{j,r}^{(t)}$ & \textbf{Feature Correlation}: $\max_{l} \langle w_{j,l,r}^{(t)}, \Mb_j^r \rangle_+$, alignment of weights to true features. \\
$d_{j,r}(\mathcal{D})$ & \textbf{Data-dependent Signal Strength}: Inherent learnability of class $j$ in modality $r$. \\
$I(Y; \Xb^r)$ & Mutual information between the labels and the features of modality $r$. \\
\bottomrule
\end{tabular}%
}
\end{table}

\section{Theoretical Framework for Modality Competition}
\label{sec:formulation}
This section establishes the theoretical foundation of our two-stage training paradigm, analyzing it as a principled approach to prevent modality competition. We first formalize the framework and introduce Effective Competitive Strength (ECS) to quantify the dynamics of competition (Section~\ref{sec:two_stage_formulation}~\ref{sec:ecs_formulation}). We then analyze the critical impact of Stage I, identifying a taxonomy of four distinct outcomes (Section~\ref{sec:four_outcome}). Finally, to bridge our theory to a practical diagnostic in real-world deep neural networks, we prove that mutual information is a measurable proxy for the ECS (Section~\ref{sec:ecs_to_mi}), which provides the cornerstone for the FastPID-guided methodology introduced in Section~\ref{sec:methods}.

\subsection{Two-Stage Training Formulation}
\label{sec:two_stage_formulation}

We introduce a structured two-stage paradigm at the core of our framework, consisting of \textbf{Stage I (unimodal training)} and \textbf{Stage II (fusion and joint training)}. This structure, while common in practice (often as a straightforward ``warm-up''), has lacked a rigorous theoretical treatment regarding its impact on the initial conditions for fusion.
Our analytical approach builds upon the work of Huang et al.~\cite{huang2022modality}, who theoretically analyzed modality competition in a single-stage joint training process(Stage II) and with random initialization.  \textit{However, different from it}, we proposed a two-stage framework where a dedicated Stage I can be strategically employed to deterministically shape the initial state for the subsequent joint fusion (Stage II), thereby preventing the competition dynamics identified in ~\cite{huang2022modality} before arising.

We formalize this process by considering a network composed of two modality-specific encoders, $\varphi_{\mathcal{M}r}$, and a shared classifier, $\mathcal{C}$, which are optimized using a cross-entropy loss, $\ell(\cdot, \cdot)$. 

In Stage 1, each encoder is trained independently with the classifier on its respective dataset $\mathcal{D}_r = \{(\Xb^r_i, y_i)\}_{i=1}^n$ for $T_{\text{pre}}^{(r)}$ iterations with the uni-modal loss $\mathcal{L}_r(\varphi_{\mathcal{M}r}) = \frac{1}{n}\sum_{i=1}^n \ell(\mathcal{C}(\varphi_{\mathcal{M}r}(\Xb^r_i)), y_i)$, producing the weight configurations, $w_{j,l,r}^{(T_{\text{pre}}^{(r)})}$.

Then, in Stage 2, both encoders are jointly optimized for $T_{\text{joint}}$ iterations, starting from their Stage 1 initializations. During this stage, features from both encoders are fused via element-wise summation and passed to the classifier, minimizing the multi-modal loss $\mathcal{L}_{\text{joint}} = \frac{1}{n}\sum_{i=1}^n \ell(\mathcal{C}(\varphi_{\mathcal{M}1}(\Xb^1_i) + \varphi_{\mathcal{M}2}(\Xb^2_i)), y_i)$. 

To analyze the learning dynamics within this two-stage framework, we adopt the sparse coding setting of~\cite{huang2022modality}. In this model, a sample $(\Xb^1, \Xb^2)$ with label $y \in [K]$ is generated via the model $\Xb^r = \Mb^r z^r + \xi^r$. For each modality $r \in [2]$, $\Mb^r \in \mathbb{R}^{d_r \times K}$ is a unitary feature dictionary, $z^r \in \mathbb{R}^K$ is a sparse code, and $\xi^r$ is noise. The sparse code $z^r$ is drawn from a mixture distribution $P_z$ with two data types based on signal strength: the sufficient class if its code $z^r$ has a strong signal for the true label $y$ (a large $z^r_y$), and the insufficient class if it is weak. The encoders $\varphi_{\mathcal{M}r}: \mathbb{R}^{d_r} \to \mathbb{R}^M$ are single-layer networks with a smoothed ReLU activation $\sigma(\cdot)$, parameterized by an integer $q \geq 3$ and a threshold $\beta=1/\polylog(K)$. The number of neurons is $M = K \cdot m$, with $m = \polylog(K)$, and the classifier $\mathcal{C}: \mathbb{R}^M \to \mathbb{R}^K$ is a single linear layer. More details are shown in the appendix.

\subsection{Formalizing Modality Competition with Effective Competitive Strength}
\label{sec:ecs_formulation}

To understand the dynamics governing modality competition at the beginning of Stage II, we must first quantify each modality's competitive potential at this critical initial state. Grounded in the data model introduced in Section~\ref{sec:two_stage_formulation}, we formalize the concept of Effective Competitive Strength defined as follows, with details provided in the supplementary material.

\begin{definition}[Effective Competitive Strength]
The \textbf{competitive strength} of a modality $r$ for learning features of class $j$ at iteration $t$ is captured by its Effective Competitive Strength (ECS), denoted $\Lambda_{j,r}^{(t)}$. It is defined as the product of two components:
\begin{equation}
\label{eq:ecs_main}
\Lambda_{j,r}^{(t)} = \underbrace{\max_{l} \langle w_{j,l,r}^{(t)}, \Mb_j^r \rangle_+}_{\text{Weight Alignment}} \cdot \underbrace{\left( d_{j,r}(\mathcal{D}) \right)^{\frac{1}{q-2}}}_{\text{Data Signal Influence}}
\end{equation}
\end{definition}

The two components of ECS are defined as follows:
\begin{enumerate}
    \item \textbf{Weight Alignment}: The term $\max_{l} \langle w_{j,l,r}^{(t)}, \Mb_j^r \rangle_+$ measures how well the current neuron weights ($w_{j,l,r}^{(t)}$) are aligned with the true feature direction for class $j$ in modality $r$ ($\Mb_j^r$). A higher value indicates that some neurons are already oriented to detect the relevant feature.

    \item \textbf{Data Signal Influence}: This term is derived from the modality's intrinsic signal strength. It is defined as:
    \begin{equation}
    \label{eq:d_signal_strength}
    d_{j,r}(\mathcal{D}) := \frac{1}{n\beta^{q-1}}\sum_ {(\Xb,y)\in\mathcal{D}_{s}}\mathbb{I}\{y=j\} (z^{r}_{j})^{q}
    \end{equation}
    This term quantifies the inherent learnability of the modality's data for a specific class. It aggregates the feature coefficients ($z^{r}_{j}$) from 'sufficient' data samples ($\mathcal{D}_{s}$), modulated by parameters of the network's smoothed ReLU activation function ($\beta, q$) introduced above. A larger $z^{r}_{j}$ leads to a higher $d_{j,r}(\mathcal{D})$ and thus a greater competitive strength.
\end{enumerate}

The significance of ECS lies in its role as the key determinant of the learning dynamics precisely at the onset of Stage II. Prior work~\cite{huang2022modality} established that competition is governed by the tensor power bound~\cite{an2017analyzing}, which dictates that a small initial imbalance in ECS is non-linearly amplified during joint training. Formally, modality $r$ is suppressed if at the start of joint training, its ECS is sufficiently smaller than its counterpart's:
\begin{equation}
\label{eq:competition_trigger_ecs}
\Lambda_{j,r}^{(0)} \cdot (1 + \beta) \leq \Lambda_{j,3-r}^{(0)}
\end{equation}
This highlights that the competition is predetermined by the \textit{initial} ECS values, $\Lambda_{j,r}^{(0)}$, at the beginning of Stage II. The central question we will address next is how to strategically shape this initial state.

\begin{figure}[tp]
    \centering
    \includegraphics[width=0.99\linewidth]{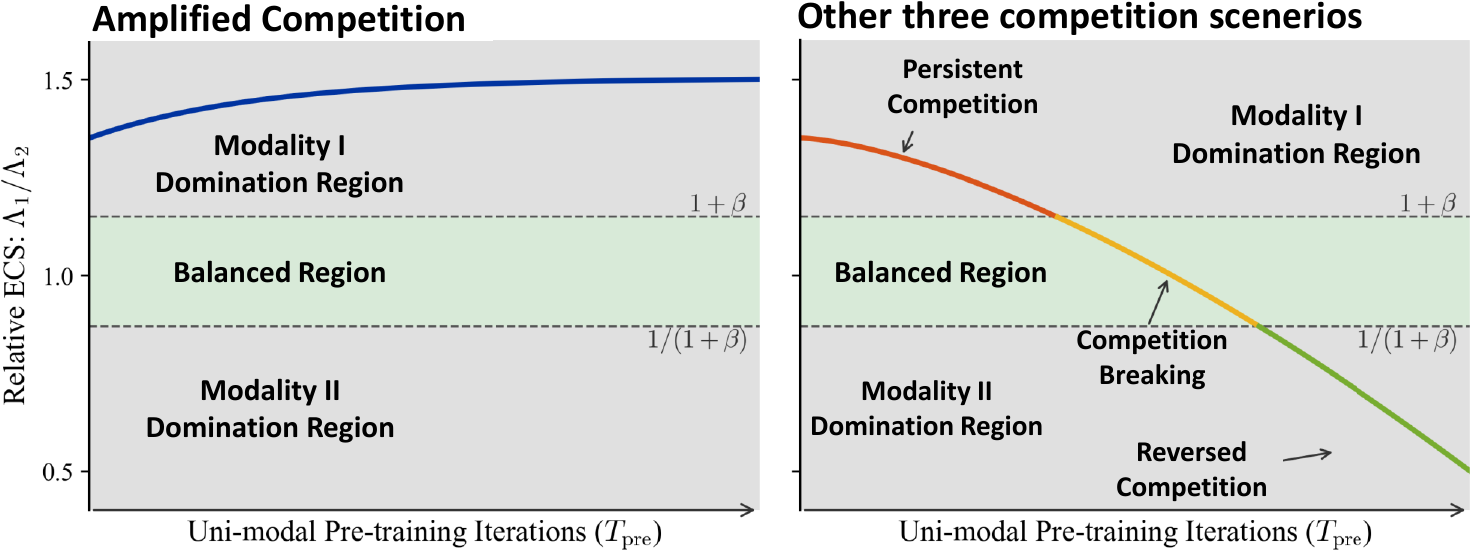}
    \caption{The illustration of the four scenerios.}
    \label{fig:four_outcome}
\end{figure}

\subsection{Impact of Stage I Unimodal Training}
\label{sec:four_outcome}

The crucial insight of our framework is that the initial ECS in the trigger condition (Eq.~\eqref{eq:competition_trigger_ecs}), $\Lambda_{j,r}^{(0)}$, is precisely the state of the encoders at the \textit{conclusion of Stage I}. This reframes the purpose of Stage I training: it is no longer a straightforward warm-up, but a deterministic process for shaping the initial states of Stage II. Therefore, the central challenge is how to strategically guide Stage I to shape the initial ECS values and deliberately set them to a state that circumvents the competition trigger, thereby enabling synergistic learning from the very start of joint training.

The following analysis reveals that this strategic shaping is a delicate process, giving rise to four distinct and predictable scenarios: three failure ones and a successful one, which we term Competition Breaking. For clarity, we will refer to the modality that starts with a natural disadvantage (i.e., a lower initial ECS, $\Lambda_{j,r}^{(0)}$) as the 'losing' modality ($r$).

\begin{enumerate}
    \item \textbf{Amplified Competition:} Training the already dominant modality further amplifies the initial imbalance, leading to more severe competition.
    \item \textbf{Persistent Competition:} Insufficient training on the weaker modality fails to break the imbalance, causing the initial competition to persist.
    \item \textbf{Reversed Competition:} Excessive training on the weaker modality overcorrects the imbalance, inverting the competition and causing the once-weaker modality to dominate.
    \item \textbf{Competition Breaking:} A precisely calibrated training duration on the weaker modality achieves a balanced initial state, nullifying the competition trigger and enabling synergistic learning.
\end{enumerate}

We now formalize these four scenarios, illustrated in Figure~\ref{fig:four_outcome}.

\subsubsection{Failure Scenarios: Amplified, Persistent, and Reversed Competition}
Incorrectly applying Stage I training fails to solve the imbalance or makes it worse. The three failure scenarios can be formally stated as the following lemmas. We provide their detailed proofs in the supplementary material.

\begin{lemma}[Amplified Competition]
When the naturally winning modality $3-r$ is trained during Stage I for any $T_{\text{pre}} > 0$, its initial ECS for joint training, $\Lambda_{j,3-r}^{(T_{\text{pre}})}$, increases, strengthening the premise of the tensor power bound and resulting in more severe competition.
\end{lemma}

\begin{lemma}[Persistent Competition]
When the losing modality $r$ undergoes a limited number of Stage I training iterations, the resulting ECS, $\Lambda_{j,r}^{(T_{\text{pre}})}$, remains insufficient to overcome the initial imbalance, and the condition in Eq.~\eqref{eq:competition_trigger_ecs} persists.
\end{lemma}

\begin{lemma}[Reversed Competition]
After an excessive number of uni-modal training iterations, the ECS of the originally losing modality $r$ grows substantially, leading to a new initial condition that reverses the original inequality:
\[
\Lambda_{j,r}^{(T_{\text{pre}})} \;\geq\; \Lambda_{j,3-r}^{(0)} \cdot (1 + \beta).
\]
This reversal causes modality $r$ to dominate, creating a new competition.
\end{lemma}

\subsubsection{Competition Breaking and Its Impact on Test Error}

\begin{lemma}[Competition Breaking]
With a carefully calibrated number of Stage I unimodal training iterations $T_{\text{pre}}$ on the losing modality $r$, the initial ECS of both modalities can be brought into near-equilibrium. This state is formally defined by the condition:
\begin{equation}
\frac{1}{1+\beta} < \frac{\Lambda_{j,3-r}^{(0)}}{\Lambda_{j,r}^{(T_{\text{pre}})}} < 1 + \beta
\end{equation}
This two-sided condition ensures that the ECS ratio falls within a range where neither modality holds a significant initial advantage, preventing the winner-takes-all dynamic. The proof is detailed in the supplementary material.
\end{lemma}

Achieving this balanced state directly improves generalization, as formalized in the following theorems.

\begin{theorem}[Test Error under Modality Competition~\cite{huang2022modality}]
In naive joint training where competition occurs, the test error is high, bounded by $\Omega \left(\frac{1}{K}\right)$, as the model relies on a single, potentially insufficient, modality.
\end{theorem}

\begin{theorem}[Test Error under Balanced Learning]
If the initial conditions are set according to the Competition Breaking lemma, both modalities learn effectively. This leads to a quadratically lower test error bound:
\begin{equation}
\text{Error}_{\text{test}}^{\text{balanced}} \leq O\left(\frac{1}{K^2}\right)
\end{equation}
\end{theorem}
This quadratic improvement demonstrates that deterministically nullifying the tensor power bound through a principled two-stage training strategy leads to significantly better generalization. The proof is detailed in the supplementary material.

\subsection{Mutual Information as a computationally tractable proxy for ECS}
\label{sec:ecs_to_mi}

The previous analysis establishes that modality competition is governed by the relative ECS at the start of joint training. However, ECS is an analytical quantity derived from a simplified model, making it intractable to measure in complex deep neural networks. This creates a significant gap between our theoretical understanding and computational implementation.

To bridge this gap, we require a measurable quantity that can serve as a faithful proxy for the uncomputable ECS. We posit that the mutual information, $I(Y; \Xb^r)$, is such a quantity. The derivation proceeds by constructing a chain of inequalities that connects the internal state of the model to its information-theoretic output. The first crucial step is to establish how the model's internal weight alignment for a single data point translates into classification confidence, which we formalize in the following lemma.

\begin{lemma}[From Feature Correlation to Conditional Entropy]
\label{lem:correlation_to_entropy}
For a sufficient data sample $(\Xb, y)$, a higher feature correlation $\Gamma_{y,r}$ for the true class leads to a larger classification margin $\Delta(\Xb^r)$, which in turn places a tighter upper bound on the model's point-wise uncertainty, as quantified by the conditional entropy $H(Y|\Xb^r = \xb^r)$:
\begin{equation}
H(Y|\Xb^r = \xb^r) \le (K-1)e^{-\Delta(\xb^r)} + O(e^{-2\Delta(\xb^r)})
\end{equation}
This establishes a direct link between a well-aligned model and low uncertainty at the sample level.
\end{lemma}

This lemma establishes the crucial link at the sample level. The next step in the derivation, detailed in the supplementary material, aggregates this effect across the entire dataset. By taking the expectation of the conditional entropy over all samples, we can derive a lower bound on the mutual information $I(Y;\Xb^r)$ in terms of the aggregated feature correlations. The final step connects this result to ECS by substituting its definition, culminating in the following theorem.

\begin{theorem}[Mutual Information as a Proxy for Competitive Strength]
\label{thm:mi_ecs_proxy}
Under the global assumptions of Section 3.1, there exists a positive data-dependent constant $\tilde{c}_r$ such that the mutual information $I(Y;\Xb^r)$ is lower-bounded by the sum of the effective competitive strengths:
\begin{align}
    I(Y;\Xb^r) &\ge \tilde{c}_r \sum_{j=1}^{K} (\ECS_{j,r})^q + O(\sigma_0+\sigma_g+\sigma_0^{q+1})
\end{align}
where
\begin{align}
    \tilde{c}_r &= \frac{K-1}{K \cdot \beta^{q-1} \cdot q} \left[\bar{d}_r(\mathcal{D})\right]^{-\frac{q}{q-2}} > 0
\end{align}
and $\bar{d}_r(\mathcal{D})$ is the average data-dependent signal strength for modality $r$.
\end{theorem}

\begin{remark}
Theorem~\ref{thm:mi_ecs_proxy} provides the theoretical cornerstone for our framework. It formally establishes that a measurable, information-theoretic quantity serves as a faithful proxy for the competition dynamics governed by ECS. This insight directly motivates our methodological approach, which leverages the information-theoretic quantity to guide the training process, as detailed in Section~\ref{sec:methods}.
\end{remark}

\section{Methods}
\label{sec:methods}

\begin{figure*}[tp]
    \centering
    \includegraphics[width=0.99\linewidth]{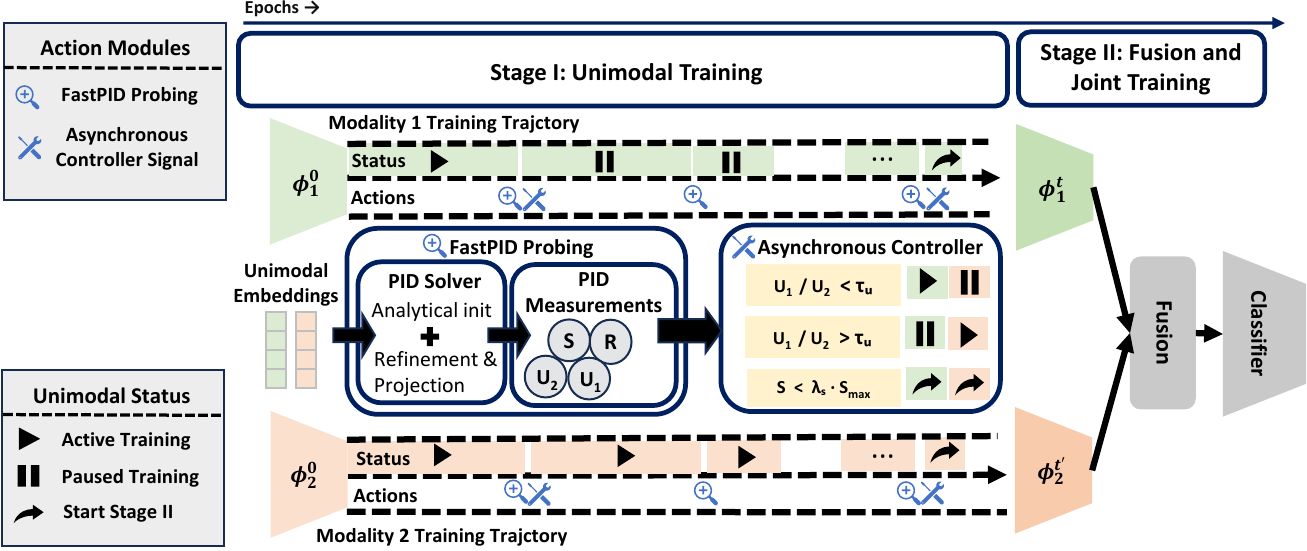}
    \caption{\textbf{Overview of our FastPID-guided two-stage asynchronous training framework.} While Stage~II follows the standard fusion-and-joint-training setup used in prior work, in Stage I, we periodically run FastPID probing, which estimates modality interactions ($U_1$, $U_2$, $R$, $S$) via a two-phase hybrid approach. Then, the asynchronous controller uses these measurements to either (i) pause the dominant modality when the uniqueness ratio exceeds a threshold, and (ii) trigger the transition to Stage II when the synergy falls. Consequently, the models enter Stage II after a different number of unimodal training epochs, denoted by the superscripts $t$ and $t'$ in $\phi_1^t$ and $\phi_2^{t'}$.}
    \label{fig:framework}
\end{figure*}

\subsection{Overall and Motivation}
\label{sec:motivation}

Our theoretical analysis (Section~\ref{sec:formulation}) establishes that Stage I training is a critical phase, whose schedule deterministically shapes the initial state for Stage II joint fusion, leading to either the three failure ones or a desirable ``Competition Breaking'' with provable lower bound. We face a fundamental challenge: \textit{How to schedule Stage I to locate the Stage II initial state without a computationally prohibitive, brute-force search over training schedules? } The core difficulty lies in the lack of a \textit{predictive diagnostic metric} as a proxy of the theoretically-derived but intractable ECS: a computable quantity available during Stage I that can offer forward-looking insight into the eventual Stage II fusion. Especially when the encoders are trained on parallel, non-communicating computational graphs in Stgae~I, the optimizers are blind to their emergent interplay before fusion. Without such a metric, the relationship between the Stage I schedule and the final performance remains a black-box optimization problem.  

To address this challenge, we propose a two-stage scheduling training framework. As illustrated in Figure~\ref{fig:framework}, it is built upon two core modules: (1) an asynchronous controller that manages the Stage I schedule by selectively training one modality's encoder while pausing the other and triggering the transition to Stage II and (2) FastPID, a computationally efficient and differentiable solver for Partial Information Decomposition (PID) that serves as the real-time diagnostic engine for the controller. 

While our theory in Section~\ref{sec:ecs_to_mi} proves that mutual information, $I(Y;X^r)$, serves as a faithful proxy, both of them are calculated from the coarse-grained marginal distributions $p(X_r, y)$, which are not accurate enough to diagnose modality competition, detailed in Section~\ref{sec:comparison_mi}. Instead, Partial Information Decomposition (PID) provides the fine-grained granularity of the joint distribution $p(X_1, X_2, Y)$ by dissecting information into interpretable atoms required by our asynchronous controller: (i) modality-specific uniqueness ($U_1, U_2$), which offers a clear signal for balancing competitive advantages, and (ii) emergent synergy ($S$), which quantifies collaborative potential to identify the ideal moment for fusion. However, applying PID as a real-time diagnostic is infeasible with conventional solvers due to their prohibitive computational cost and lack of differentiability. To bridge this gap, we proposed FastPID, a computationally efficient and differentiable solver that makes our control strategy practical. 

The following sections detail the two core components: the asynchronous controller leveraging uniqueness and synergy to steer Stage I (Section~\ref{sec:framework}), and the FastPID solver that provides these modality interaction measurements in real-time (Section~\ref{sec:fastpid}).

\begin{algorithm}[htb]
   \caption{FastPID-Guided Asynchronous Multimodal Training}
   \label{alg:pid_training}
\begin{algorithmic}[1]
   \STATE Input: Unimodal encoders $E_1, E_2$; Datasets $\mathcal{D}_1, \mathcal{D}_2$.
   \STATE Parameters: Probing frequency $f_p$, uniqueness threshold $\tau_u$, synergy threshold $\lambda_s$.
   \STATE Initialize: $current\_stage \leftarrow \texttt{unimodal}$ \\ $best\_synergy \leftarrow -1$.
   \FOR{$epoch = 1 \to max\_unimodal\_epochs$}
        \IF{$epoch \pmod{f_p} == 0$} 
            \STATE Extract embeddings from $E_1, E_2$, compute joint distribution $P(X_1, X_2, Y)$.
            \STATE Calculate $(U_1, U_2, R, S)$ using \texttt{FastPID}.

            \IF{$U_1 / (U_2 + \epsilon) > \tau_u$} 
                \STATE $active\_modality\_idx \leftarrow 2$
            \ELSIF{$U_2 / (U_1 + \epsilon) > \tau_u$} 
                \STATE $active\_modality\_idx \leftarrow 1$
            \ENDIF

            \IF{$S < \lambda_s \cdot best\_synergy$ $\AND$ $best\_synergy > 0$} 
                \STATE $current\_stage \leftarrow \texttt{joint}$
                \STATE \textbf{break} 
            \ENDIF
            \STATE $best\_synergy \leftarrow \max(best\_synergy, S)$.
        \ENDIF
        \STATE Train encoder $E_{active\_modality\_idx}$ on its dataset $\mathcal{D}_{active\_modality\_idx}$.
   \ENDFOR
   \STATE
   \STATE Unfreeze both encoders $E_1, E_2$ and start joint fusion training.
\end{algorithmic}
\end{algorithm}

\subsection{Asynchronous Controller}
\label{sec:framework}
Illustrated in Figure~\ref{fig:framework}, we propose an asynchronous training controller to schedule the Stage I training and locate the ideal Stage II initial state, leveraging the PID interactions computed by \textit{FastPID}. Instead of a brute-force search, it periodically assesses the information state of the unimodal encoders and uses this diagnosis to control the unimodal training schedule. Our controller addresses the challenge of navigating the complex and non-monotonic information landscape revealed in our analysis (Section~\ref{sec:analysis}). 

The controller works under two principled selection criteria:
\begin{enumerate}
    \item \textbf{Constraint on Modality Uniqueness:} To prevent one modality from dominating, the controller monitors the balance of unique information ($U_1, U_2$). If their ratio exceeds a threshold $\tau_u$, the training of the more dominant modality is paused to allow the weaker one to ``catch up.''
    \item \textbf{Ideal Initial State Detection:} The controller tracks synergy ($S$). It switches to Stage II joint training once it detects that $S$ has peaked and begun to decline (e.g., dips below a fraction $\lambda_s$ of its maximum observed value), indicating the degradation of the potential for synergistic fusion.
\end{enumerate}

The details of this dynamic control strategy are formalized in Algorithm~\ref{alg:pid_training}. The control operates particularly in Stage I by periodically probing the model state at a frequency $f_p$. At each probe, the unimodal encoders are frozen and their embeddings are used to compute the PID measures via \textit{FastPID}. These measures immediately inform the control logic. First, the uniqueness ratio is checked against $\tau_u$ to enforce modality balance by pausing the dominant modality and starting the other one. 
Second, inspired by our analysis in Section~\ref{sec:analysis} that synergistic potential is non-monotonic as Stage I training continues, the controller tracks the synergy ($S$) and transitions to Stage II once $S$ shows a meaningful decline from this peak.
This process transforms the search for an ideal starting point from a manual, brute-force problem into an automated and adaptive procedure, ensuring a balanced and synergistic state for joint training.

\subsection{FastPID: A Fast and Accurate Hybrid PID Solver}
\label{sec:fastpid}
The effectiveness of the asynchronous controller introduced in Section~\ref{sec:framework} depends on the ability to obtain accurate periodic, real-time modality interaction measurements of uniqueness and synergy. To enable such a practical implementation for solving PID as real-time diagnostics, we propose FastPID, an accurate and efficient computational solver.

\subsubsection{Conventional PID solver and our Hybrid Approach}
To calculate the PID measurements (introduced in Section~\ref{sec:pidbg}), previous methods~\cite{bertschinger2014quantifying,liang2024quantifying} solves a constrained convex optimization problem to find a distribution $q^*$:
\begin{equation}
q^* = \arg\max_{Q \in \Delta_p} H_Q(Y \mid X_1, X_2),
\label{eq:pid_optimization}
\end{equation}
where the feasible set $\Delta_p$ preserves the marginals from the empirical distribution $p$.
However, this is impractical for deep neural networks as it suffers from two critical bottlenecks: (1) prohibitive computational complexity, as the optimization space grows multiplicatively, and (2) a non-differentiable solution process, which prevents end-to-end integration.

FastPID overcomes bottlenecks of these PID solvers by adopting a hybrid strategy that combines the speed and stability of a closed-form analytical proxy with the precision and differentiability of iterative, gradient-based refinement.
FastPID operates in two phases: (1) an analytical initialization phase that constructs a principled, closed-form, non-synergistic proxy $Q_{\text{init}}$ that exactly satisfies the marginal constraints; and (2) a refinement phase that optimizes conditional entropy with projected gradient steps, converging toward the optimum $q^*$ under the PID objective. This design makes FastPID fully differentiable, computationally efficient, and sufficiently accurate for periodic probing and direct integration into modern gradient-based training pipelines.

\subsubsection{From Analytical Proxy to Differentiable Refinement}
We now present FastPID’s two-phase procedure in detail, which moves from a principled analytical initialization to an iterative differentiable refinement with projection to the required marginals.

\noindent
\textbf{Phase I: Analytical Initialization}
Rather than starting from a random point, FastPID begins with a principled initial approximation, $Q_{\text{init}}$. Our key insight is to derive this from a closed-form, non-synergistic baseline that assumes conditional independence: $q(x_1, x_2 \mid y) = q(x_1 \mid y)\, q(x_2 \mid y)$.

Applying this assumption along with the problem's marginal constraints ($q(x_i, y) = p(x_i, y)$ and $q(y) = p(y)$) allows us to derive the proxy directly:
\begin{align*}
q(x_1, x_2, y) &= q(y) \, q(x_1 \mid y) \, q(x_2 \mid y) \\
&= p(y) \, \frac{p(x_1, y)}{p(y)} \, \frac{p(x_2, y)}{p(y)} = \frac{p(x_1, y) p(x_2, y)}{p(y)}
\end{align*}
This yields the final, closed-form analytical proxy:
\begin{equation}
\boxed{ Q_{\text{init}}(x_1, x_2, y) = \frac{P(x_1, y)\; P(x_2, y)}{P(y)} }
\label{eq:analytical_q}
\end{equation}

This initialization provides FastPID with a computationally efficient and theoretically meaningful head start, positioning the solver much closer to the true solution than a random guess, detailed in Section~\ref{sec:pid_ini}. However, relying on such a simplified assumption may deviate from the true optimum $q^*$ in~\eqref{eq:pid_optimization}, which motivates the next phase.

\noindent
\textbf{Phase II: Differentiable Refinement with Projection}
To bridge the gap between the analytical proxy and the true optimum, we introduce a fully differentiable iterative refinement process. Instead of directly optimizing the constrained probability distribution $Q$, which is numerically challenging, we employ a re-parameterization strategy. We define an unconstrained tensor of \textit{logits}, $\boldsymbol{\theta}$, such that the distribution $Q$ can be recovered at any time via the softmax function, $Q = \text{softmax}(\boldsymbol{\theta})$. This elegantly handles the non-negativity and sum-to-one constraints, allowing us to use powerful unconstrained optimizers like Adam~\cite{kingma2014adam}.

Starting from the logits corresponding to $Q_{\text{init}}$, the refinement process directly optimizes the original objective in Equation~\eqref{eq:pid_optimization} by maximizing $H_Q(Y \mid X_1, X_2)$. Each refinement step consists of:

\begin{enumerate}[leftmargin=*, itemsep=1pt]
    \item \emph{Distribution Generation:} From the current logits $\boldsymbol{\theta}_k$, generate the probability distribution $Q_k$.
    \begin{equation}
        Q_k = \text{softmax}(\boldsymbol{\theta}_k)
    \end{equation}
    This distribution is valid (non-negative and sums to one) but does not yet satisfy the marginal constraints of the feasible set $\Delta_p$.

    \item \emph{Projection Step:} Project the generated distribution $Q_k$ back onto the feasible set $\Delta_p$ to enforce the required marginals, yielding the projected distribution $Q_{k, \text{proj}}$.
    \begin{equation}
        Q_{k, \text{proj}} = \text{Project}_{\Delta_p}(Q_k) = \arg\min_{Q' \in \Delta_p} D(Q' \Vert Q_k)
    \end{equation}
    where $D$ is a divergence metric. This projection is efficiently implemented using an iterative scaling method akin to Sinkhorn-Knopp~\cite{cuturi2013lightspeed}, which alternately normalizes the distribution to match the empirical marginals $p(x_1, y)$ and $p(x_2, y)$.

    \item \emph{Loss Calculation and Gradient Update:} The loss, $\mathcal{L}$, is calculated as the negative conditional entropy of the \textit{projected} distribution, $\mathcal{L}(Q_{k, \text{proj}}) = -H_{Q_{k, \text{proj}}}(Y \mid X_1, X_2)$. Crucially, the gradient of this loss is backpropagated through the projection step to the original logits $\boldsymbol{\theta}_k$. The Adam optimizer then uses this gradient to compute the next iterate of the logits:
    \begin{equation}
        \boldsymbol{\theta}_{k+1} = \text{Adam}(\boldsymbol{\theta}_k, \nabla_{\boldsymbol{\theta}_k} \mathcal{L}(Q_{k, \text{proj}}))
    \end{equation}
\end{enumerate}

To summarize, FastPID's two-phase procedure is unified as follows. Phase I provides a principled initialization by setting the initial logits to $\boldsymbol{\theta}_0 = \log Q_{init}$, using the analytical proxy from~\eqref{eq:analytical_q}. Notably, since $Q_{\text{init}}$ is constructed to satisfy the marginal constraints, this starting point already lies within the feasible set $\Delta_p$.

Phase II then performs iterative refinement on these logits. Each step follows the gradient ascent update rule:

\begin{align}
    \label{eq:fastpid_update}
    \boldsymbol{\theta}_{k+1} = \\
    \text{Adam} \bigg( & \boldsymbol{\theta}_k, 
    & \nabla_{\boldsymbol{\theta}} \left[ H_{\Pi_{\Delta_p}(\text{softmax}(\boldsymbol{\theta}))}(Y \mid X_1, X_2) \right] \bigg|_{\boldsymbol{\theta}=\boldsymbol{\theta}_k} \bigg) \nonumber
\end{align}
This approach unifies the two phases into a simple yet powerful paradigm: initialize within the feasible set, then optimize in an unconstrained space via a differentiable projection to solve the maximization goal in~\eqref{eq:pid_optimization}. The number of refinement steps is a tunable hyperparameter, balancing computational speed and approximation accuracy.


\begin{algorithm}[h!]
\caption{The FastPID Algorithm}
\label{alg:fastpid}
\begin{algorithmic}[1]
\STATE \textbf{Input}: Joint distribution $p(x_1, x_2, y)$, 
STATE Parameters: max iterations $K$, learning rate $lr$, tolerance $tol$.

\STATE \textbf{Initialize:}
\STATE Calculate analytical proxy: $Q_{\text{init}} \leftarrow \frac{p(x_1, y)\, p(x_2, y)}{p(y)}$
\STATE Initialize optimizable logits: $\boldsymbol{\theta}_0 \leftarrow \log(Q_{\text{init}})$
\STATE Initialize optimizer: $\text{Optimizer} \leftarrow \text{Adam}(\text{params}=[\boldsymbol{\theta}_0], \text{lr}=lr)$
\STATE $Q_{\text{old}} \leftarrow Q_{\text{init}}$

\STATE \textbf{Refine:}
\FOR{$k = 0$ \TO $K-1$}
    \STATE $\text{Optimizer.zero\_grad}()$
    \STATE Generate current distribution: $Q_k \leftarrow \text{softmax}(\boldsymbol{\theta}_k)$
    \STATE Project to feasible set: $Q_{k, \text{proj}} \leftarrow \text{Project}_{\Delta_p}(Q_k)$ \quad \COMMENT{via Sinkhorn scaling}
    \STATE Calculate loss: $\mathcal{L} \leftarrow -H_{Q_{k, \text{proj}}}(Y \mid X_1, X_2)$
    \STATE $\mathcal{L}.\text{backward}()$ \quad \COMMENT{Computes $\nabla_{\boldsymbol{\theta}_k} \mathcal{L}$}
    \STATE $\text{Optimizer.step}()$ \quad \COMMENT{Updates $\boldsymbol{\theta}_k \to \boldsymbol{\theta}_{k+1}$}
    
    \IF{$\max(|Q_{k, \text{proj}} - Q_{\text{old}}|) < tol$}
        \STATE \textbf{break}
    \ENDIF
    \STATE $Q_{\text{old}} \leftarrow Q_{k, \text{proj}}$
\ENDFOR

\STATE \textbf{Finalize:}
\STATE Get final optimized distribution: $Q^* \leftarrow \text{Project}_{\Delta_p}(\text{softmax}(\boldsymbol{\theta}_{K}))$

\STATE \textbf{Calculate PID Terms and return} $S, R, U_1, U_2$
\end{algorithmic}
\end{algorithm}

\subsubsection{The Complete FastPID Algorithm}
The FastPID algorithm is summarized in Algorithm~\ref{alg:fastpid}. It first computes the analytical initialization $Q_{\text{init}}$, then iteratively refines it, and finally calculates the PID terms using the refined distribution $Q_{\text{final}}$. The entire pipeline is implemented with differentiable tensor operations, making it suitable for end-to-end training.

\section{Experiment and Results}

\begin{table*}[htp]
\centering
\caption{Performance comparison with state-of-the-art methods on several mainstream multimodal learning benchmarks. Our method demonstrates competitive performance across datasets with diverse characteristics, including AVE, CREMA-D, CGMNIST, UCF-101. }
\label{tab:performance_ave_crema_cgmnist}
\begin{subtable}{.35\textwidth} 
    \centering
    \caption{CREMA-D and AVE (\%)}
    \label{tab:comparison_crema_ave_v2} 
    \begin{tabular}{ l|cc}
    \toprule
    Method & CREMA-D & AVE \\
    \midrule
    Unimodal1 & 56.67 & 59.37 \\
    Unimodal2 & 50.14 & 30.46 \\
    \midrule
    Concat Fusion & 59.50 & 62.68 \\
    G-Blending~\cite{wang2020makes} & 63.81 & 62.75 \\
    OGM-GE\cite{peng2022balanced} & 65.59 & 62.93 \\
    PMR\cite{fanPMRPrototypicalModal2023} & 66.10 & 64.20 \\
    UMT\cite{du2023uni} & 70.97 & 67.71 \\
    MMPareto~\cite{weiMMParetoBoostingMultimodal2024} & 75.13 & 68.90 \\
    DiagRe~\cite{wei2024diagnosing} & 72.31 & 72.18 \\
    ReconBoost\cite{hua2024reconboost} & 79.82 & 71.35 \\
    \midrule
    Ours  & \textbf{83.50} & \textbf{73.63} \\
    \bottomrule
    \end{tabular}
\end{subtable}%
\hfill
\begin{subtable}{.25\textwidth} 
    \centering
    \caption{CGMNIST (\%)}
    \label{tab:comparison_cgmnist_v2} 
    \begin{tabular}{l|c}
    \toprule
    Method & CGMNIST \\
    \midrule
    Unimodal1 & 99.30 \\
    Unimodal2 & 60.40 \\
    \midrule
    Concat Fusion & 53.30 \\
    Conv Pareto~\cite{sener2018multi} & 62.00\\
    GradNorm~\cite{chen2018gradnorm} & 76.16 \\
    PCGrad~\cite{yu2020gradient}  & 79.35 \\
    Uniform~\cite{weiMMParetoBoostingMultimodal2024} & 75.68 \\
    MetaBalance~\cite{he2022metabalance}  & 79.18 \\
    PMR~\cite{fanPMRPrototypicalModal2023} & 74.20 \\
    MMPareto~\cite{weiMMParetoBoostingMultimodal2024} & 81.88 \\
    \midrule
    Ours & \textbf{95.79} \\
    \bottomrule
    \end{tabular}
\end{subtable}%
\hfill
\begin{subtable}{.35\textwidth} 
    \centering
    \caption{UCF-101 (\%)} 
    \label{tab:comparison_ucf101_v2} 
    \begin{tabular}{l|c}  
    \toprule
    Method & UCF-101 \\
    \midrule
    Unimodal1 & 70.55 \\
    Unimodal2 & 78.60 \\
    \midrule
    Concat Fusion &  81.80 \\
    G-Blending~\cite{wang2020makes} & 68.82 \\
    OGM-GE\cite{peng2022balanced} & 82.07 \\
    UMT\cite{du2023uni} & 84.18 \\
    PMR\cite{fanPMRPrototypicalModal2023} &  81.93 \\
    ReconBoost\cite{hua2024reconboost} & 82.89 \\
    MMPareto~\cite{weiMMParetoBoostingMultimodal2024} & 85.30 \\
    DiagRe~\cite{wei2024diagnosing} & 82.87 \\
    \midrule
    Ours & \textbf{93.78} \\
    \bottomrule
    \end{tabular}
\end{subtable}
\end{table*}

\subsection{Datasets}
Following existing literature and formal evaluation protocols~\cite{hua2024reconboost, weiMMParetoBoostingMultimodal2024, wei2024diagnosing}, we evaluate our framework on a diverse set of datasets spanning multiple domains, including audio-visual classification, image-text classification, and action recognition, ensuring the generalization of our approach across different modalities and tasks.

\noindent
\textbf{CREMA-D}~\cite{cao2014crema} is a classic multi-modal emotion recognition dataset featuring synchronized facial expressions and vocal recordings. It contains 7,442 clips from 91 actors, expressing 6 distinct emotional states. We randomly split the dataset into 6,698 training samples and 744 testing samples.

\noindent
\textbf{AVE}~\cite{tian2018audio} represents a more challenging scenario with complex real-world audio-visual events. It contains 4,143 10-second YouTube videos spanning 28 event classes, with both visual scenes and ambient sounds. Unlike CREMA-D's controlled environment, AVE covers a wider range of domains, including human activities, animal activities, music performances, etc.

\noindent
\textbf{Colored-and-gray-MNIST}~\cite{kim2019learning} serves as a controlled synthetic benchmark specifically designed to study modality reliability. Based on MNIST~\cite{lecun1998gradient}, each instance pairs a grayscale digit image with a monochromatic colored version. This controlled setting allows us to quantitatively evaluate how our method handles known modality biases.

\noindent
\textbf{UCF-101}~\cite{soomro2012ucf101} is an action recognition dataset with two modalities, RGB and optical flow. This dataset contains 101 categories of human actions. The two modalities provide complementary information, with RGB capturing static appearance and optical flow capturing motion dynamics. The entire dataset is divided into a 9,537-sample training set and a 3,783-sample test set according to the original setting.

\subsection{Implementation Details}
Unless specified otherwise, we adopt ResNet-18~\cite{he2016deep} as the encoder backbone. To fairly assess learning dynamics, encoders are trained from scratch for all datasets. We adapt the backbone for several data types: the audio encoder's input layer is modified for single-channel spectrograms, and for optical flow, horizontal and vertical fields are stacked into a two-channel input. When processing visual inputs from multiple frames, the frames are stacked along the channel dimension and fed into the 2D network as a multi-channel input. The exception is the CG-MNIST benchmark, for which we employ a lightweight 4-layer CNN as~\cite{fanPMRPrototypicalModal2023}. Input data is prepared as follows: all audio is re-sampled to 16kHz and converted into spectrograms via a 512-point STFT (e.g., yielding a 257$\times$299 map for CREMA-D). Visual inputs are formed by deterministic frame sampling from each video. For instance, a single frame is randomly sampled for CREMA-D, while for AVE, three frames are sampled uniformly from the available frames.

Our training follows a two-stage process: FastPID-guided unimodal training (up to 150 epochs) followed by joint training (100 epochs). We use an SGD optimizer~\cite{robbins1951stochastic} with 0.9 momentum, 1e-4 weight decay, and a batch size of 64. The learning rate was initialized based on the dataset: 1e-3 for the CREMA-D and UCF-101 datasets, and 1e-2 for the CGMNIST and AVE datasets. For all setups, the rate was reduced by a factor of 10 every 30 epochs via a StepLR scheduler. During the Stage~II training, we follow the uniform baseline from ~\cite{weiMMParetoBoostingMultimodal2024}, where multimodal and unimodal losses are summed equally. For all PID computations, we append an MLP layer to each encoder to project features into a common 64-dimensional space, ensuring a fair comparison between conventional PID solvers and our \texttt{FastPID}.

\subsection{Comparison with State-of-the-arts}
We conduct a comprehensive evaluation of our proposed methods against a suite of established baselines and state-of-the-art methods for modality competition. The results, summarized in Table~\ref{tab:performance_ave_crema_cgmnist}, demonstrate that our approach achieves superior performance across all four benchmark datasets, which span diverse modalities and imbalance characteristics.

\noindent
\textbf{Performance on Audio-Visual Classification} On the audio-visual classification tasks, CREMA-D and AVE, our method sets a new state-of-the-art. For CREMA-D, we achieve an accuracy of 83.50\%, significantly outperforming the next best method, ReconBoost~\cite{hua2024reconboost}, by \textbf{3.68\%}. Similarly, on the more complex in-the-wild AVE dataset, our framework achieves 73.63\% accuracy, surpassing the previous best result from DiagRe~\cite{wei2024diagnosing}. These results highlight our method's efficacy in handling real-world data where the interplay between modalities is nuanced and dynamic.

\noindent
\textbf{Performance on CGMNIST and Action Recognition} Another striking result is observed on the CGMNIST benchmark, which is specifically designed to study severe modality competition. Our method achieves an accuracy of 95.79\%, a remarkable improvement of \textbf{13.91\%} over the strongest competitor, MMPareto~\cite{weiMMParetoBoostingMultimodal2024}.
Furthermore, on the UCF-101 action recognition dataset, which uses RGB and optical flow modalities, our approach reached 93.78\% accuracy, representing a substantial \textbf{8.48\%} improvement over the best-performing baseline. 

The consistent and significant performance gains across these varied domains from audio-visual and image-text to action recognition validate the generalizability and robustness of our proactive, information-theoretic approach to mitigating modality competition.

\section{Analysis} 
\label{sec:analysis}


\begin{figure*}[t!]
    \centering
    \subfloat[Uniqueness Ratio ($\tau_u$)]{%
        \includegraphics[width=0.24\textwidth]{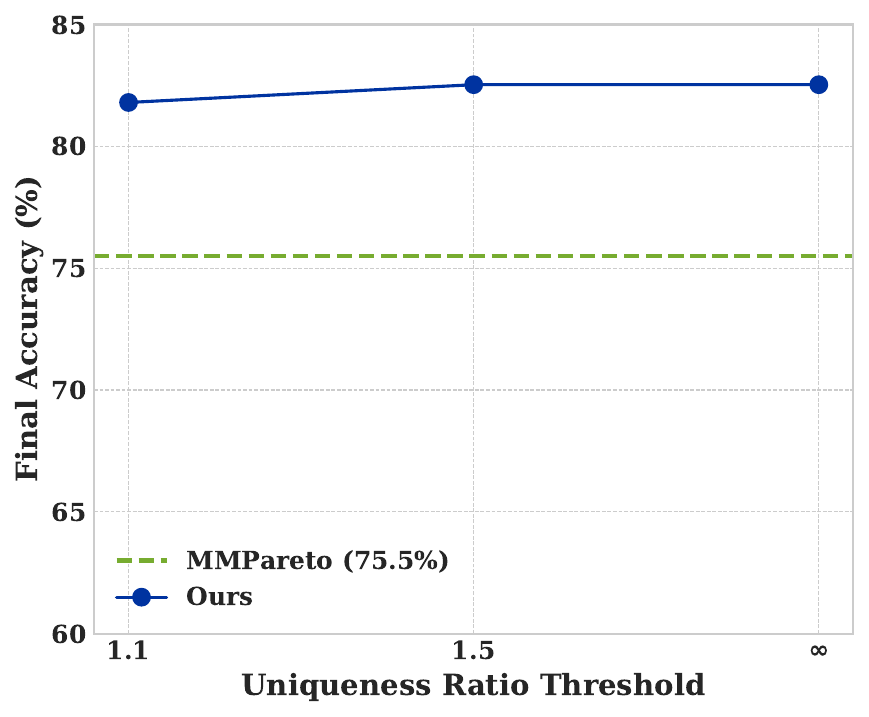}%
        \label{fig:ablation_tau}%
    }
    \hfill 
    \subfloat[Synergy Threshold ($\lambda_s$)]{%
        \includegraphics[width=0.24\textwidth]{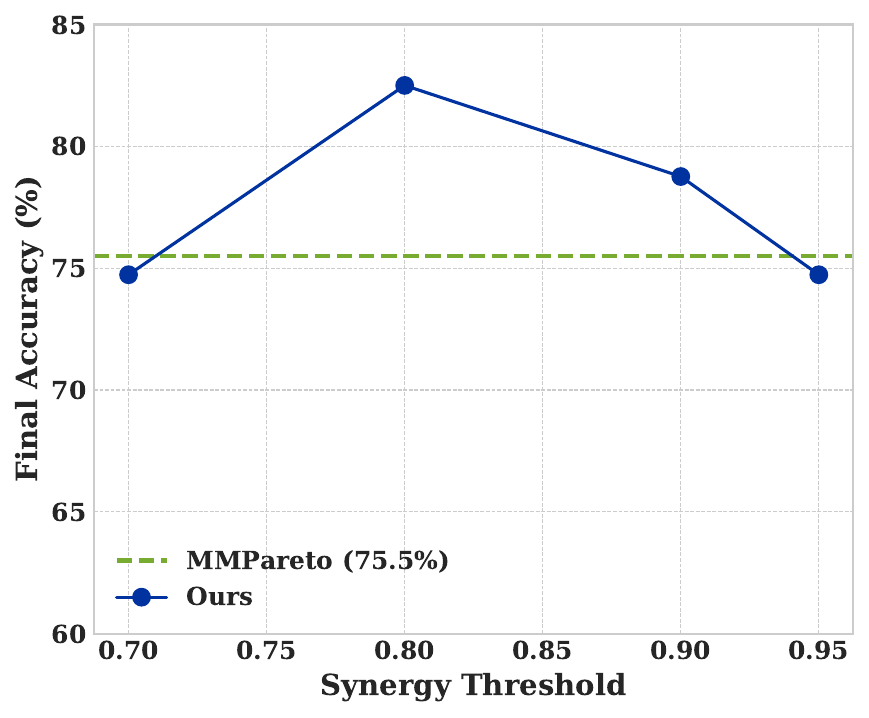}%
        \label{fig:ablation_lambda}%
    }
    \hfill 
    \subfloat[Probing Frequency ($f_p$)]{%
        \includegraphics[width=0.24\textwidth]{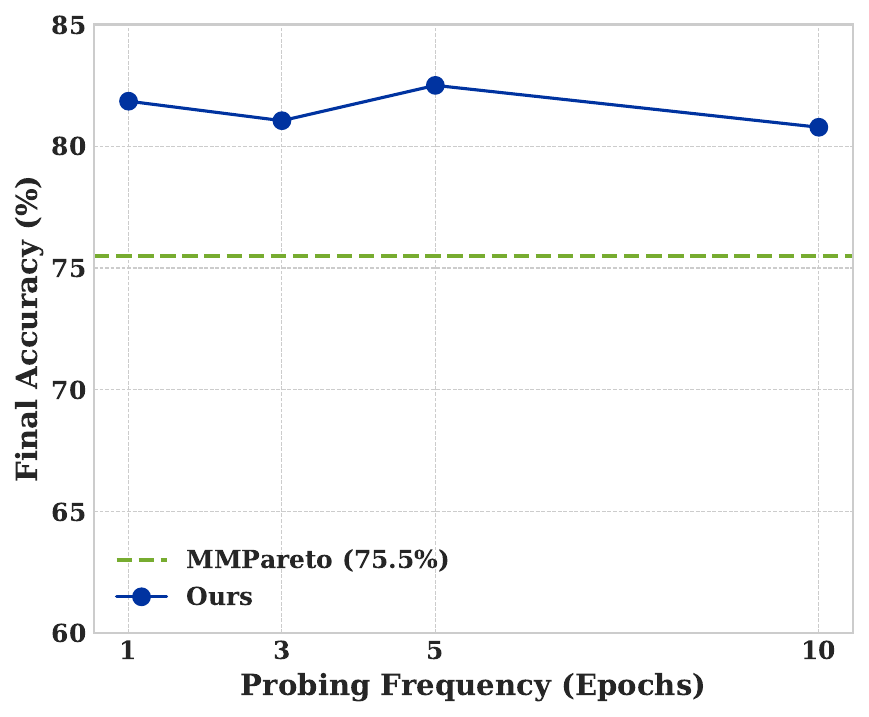}%
        \label{fig:ablation_fp}%
    }
    \hfill 
    \subfloat[Stage 2 Learning Rate]{%
        \includegraphics[width=0.24\textwidth]{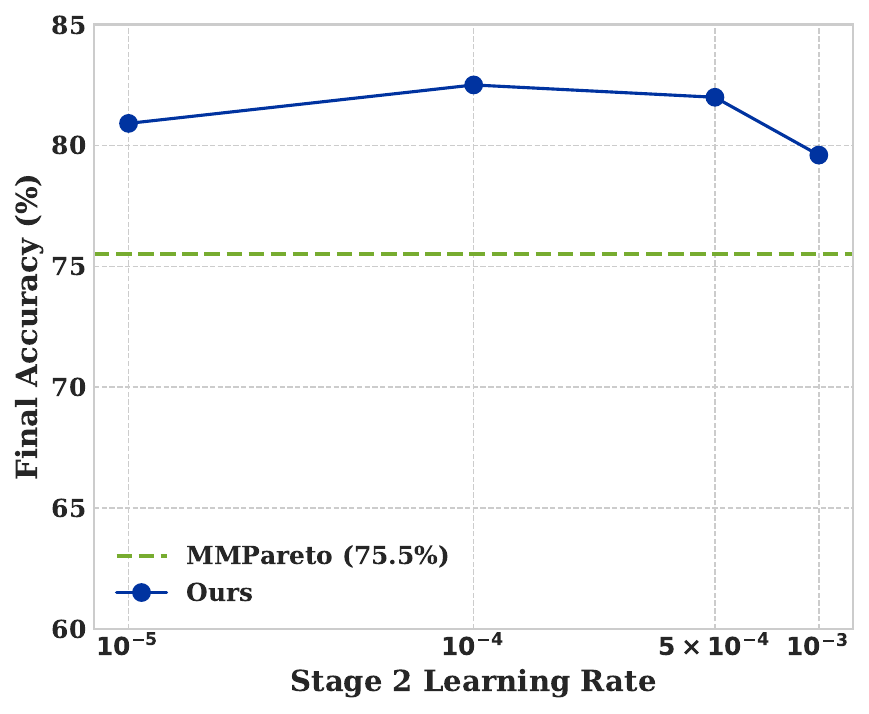}%
        \label{fig:ablation_stage2_lr}%
    }
    
    \caption{Ablation study of key framework hyperparameters and Stage 2 learning rate on the CREMA-D dataset.}
    \label{fig:ablation_hyperparams}
\end{figure*}

\subsection{Ablation Study}
We conduct a series of ablation studies on the CREMA-D dataset to dissect the effectiveness of our key components in the framework. We first analyze the sensitivity of the asynchronous controller to its core hyperparameters(Section~\label{sec:hyper}), then investigate how our FastPID-guided Stage~II initialization reshapes the optimization landscape by inspecting its responses to different Stage~II learning rate(Section~\label{sec:hyper_lr}).

\subsubsection{Framework Hyperparameter}
\label{sec:hyper}
The effectiveness of our asynchronous training framework is affected by three key hyperparameters: the uniqueness ratio threshold ($\tau_u$), the synergy threshold ($\lambda_s$), and the probing frequency ($f_p$). $\tau_u$ controls the tolerance for modality imbalance, $\lambda_s$ determines the transition point to joint training, and $f_p$ balances control granularity with computational cost. We varied each parameter while keeping others at their default values ($\tau_u=5.0$, $\lambda_s=0.95$, $f_p=5$).The results, presented in Figure~\ref{fig:ablation_hyperparams}, offer critical insights into the framework's dynamics.

For the \textbf{uniqueness ratio ($\tau_u$)}, we observe that a moderate value of 1.5 yields the best performance, which is interestingly matched by having no dynamic balancing at all ($\tau_u = \infty$). A very strict balancing threshold ($\tau_u = 1.1$) slightly degrades performance. This suggests that while our method effectively prevents runaway modality dominance, the primary driver of performance gain is not forcing perfect informational equality, but rather identifying the correct moment to transition.

This is further evidenced by the results for the \textbf{synergy threshold ($\lambda_s$)}. Performance is highly sensitive to this parameter, with a peak value at $\lambda_s = 0.80$. This indicates that the ideal transition point is not immediately after synergy peaks (e.g., $\lambda_s = 0.95$), but after allowing for a slight, principled decline. Training in Stage~I too long (a lower $\lambda_s$) or transitioning to Stage~II too early (a higher $\lambda_s$) is suboptimal, implicitly confirming the existence of a well-defined "Competition Breaking" window that our synergy-tracking mechanism successfully identifies.

Finally, the \textbf{probing frequency ($f_p$)} shows a clear trade-off between responsiveness and stability. Probing too frequently ($f_p \leq 3$) or too infrequently ($f_p = 10$) leads to lower accuracy. A frequency of 5 epochs provides the best balance, allowing the controller to make informed decisions based on stable trends rather than transient noise, without risking missing the optimal transition window.



\subsubsection{Stage II Learning Rate}
\label{sec:hyper_lr}
To comprehensively quantify the benefits of our proactive approach over standard joint training, we investigate how the altered initial state from our FastPID-guided Stage~I impacts the optimization landscape of the subsequent fusion stage by analyzing the model's performance under different learning rates during Stage~II (Figure~\ref{fig:ablation_hyperparams}). The results not only show our method's strong robustness across various learning rates, but also that its peak performance is achieved with a smaller rate of 1e-4. This contrasts with standard from-scratch training, where a larger learning rate (e.g., 1e-3) is often required for effective optimization. For our pre-conditioned model, such a large rate degrades performance (79.60\%), suggesting it overshoots the favorable loss basin established in Stage 1. This shift in the optimal learning rate provides strong evidence that our method creates a well-conditioned, easier-to-optimize landscape, making the joint training both more stable and efficient.

\subsection{Impact Analysis on Stage I Unimodal Training}

\begin{figure}[htp]
    \centering
    \includegraphics[width=0.95\linewidth]{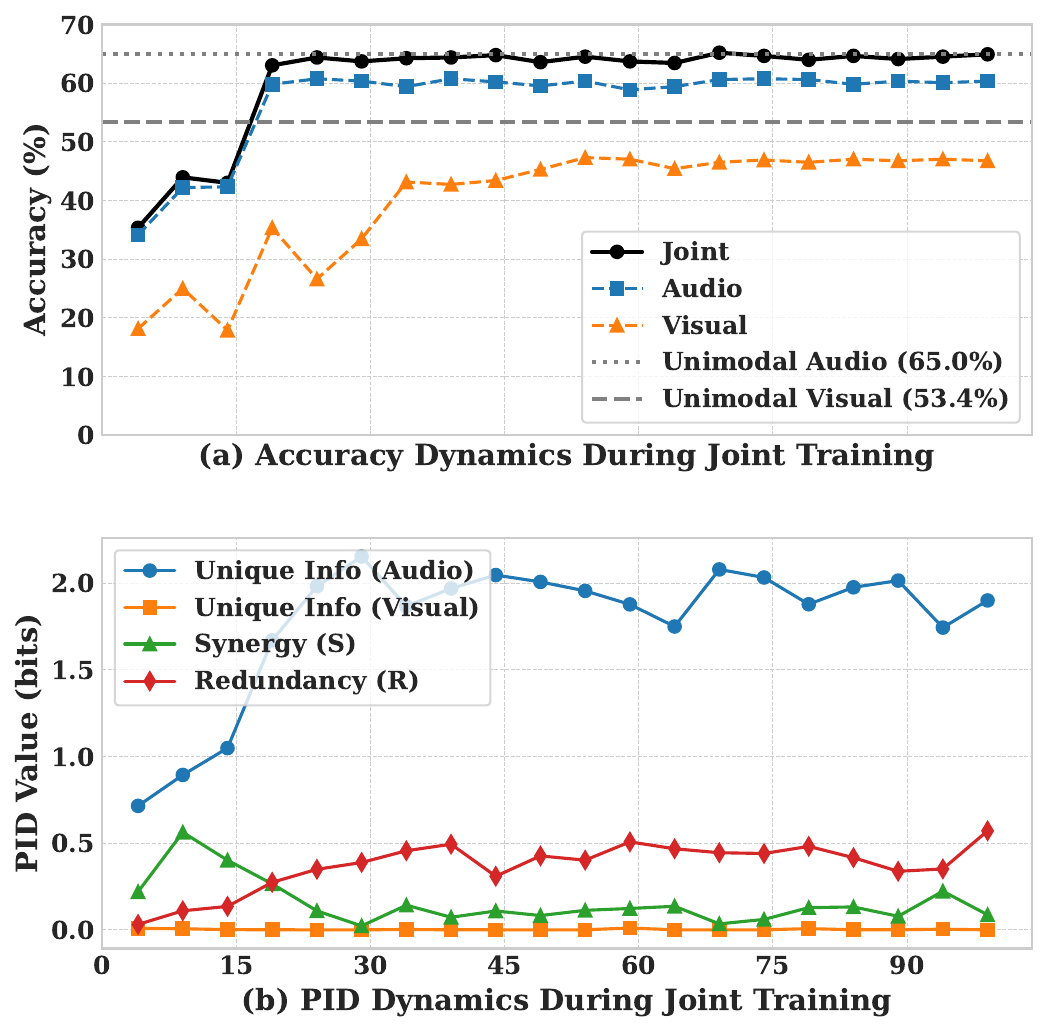}
    \caption{The PID value during the training process of the naive joint training.}
    \label{fig:pid_competition}
\end{figure}

\subsubsection{Inspect modality competition from the information perspective}
\label{sec:analysis_inspect_competition}
Quantifying modality competition presents a significant challenge because it is not a static property of a dataset but an emergent dynamic of the training process. There exists no ground-truth label for competition, but it can only be inferred indirectly and post-hoc through observing suppressed final performance or the under-optimization of one modality's branch. Relying on such lagging, coarse-grained indicators makes it difficult to diagnose and intervene before the detrimental effects become entrenched. 

To demonstrate the \textbf{diagnostic power of our framework in understanding modality competition}, we first examine a naive joint training process that just applies concatenation fusion without any balancing methods. The results are illustrated in Figure~\ref{fig:pid_competition}. 
The performance plot (Fig.~\ref{fig:pid_competition}(a)) demonstrates a severe learning imbalance: naive joint training suppresses the performance of the constituent parts compared to what they can achieve in isolation. The visual modality's accuracy stagnates below 48\%, significantly underperforming its unimodal potential of 53.4\%. Even the dominant audio modality is slightly hindered, failing to consistently reach its standalone benchmark of 65.0\% during the joint training process.

\begin{figure*}[tp]
    \centering
    \includegraphics[width=0.99\linewidth]{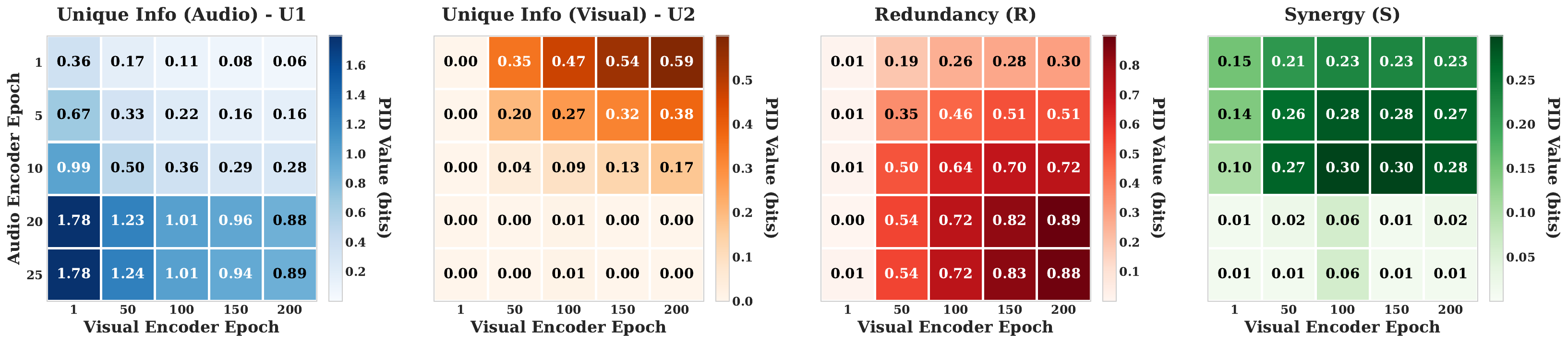}
    \caption{Grid analysis: FastPID showing the interactions at the start of joint training across a 5x5 grid of pre-trained unimodal encoder combinations.}
    \label{fig:pid_grid}
\end{figure*}

Our PID analysis (Fig.~\ref{fig:pid_competition}(b)) deconstructs this failure at an information-theoretic level. The unique information from the audio modality is overwhelmingly dominant (\(\sim\)2.0 bits), while the unique information from the visual modality is suppressed to near-zero. More critically, the synergy term~(S), which quantifies emergent cross-modal information, remains negligible, confirming that the model fails to learning collaborative representations but is instead defaulting to the stronger modality. 
This observation validates the need for our proactive, FastPID-guided framework, which is designed to prevent such competitive dynamics from arising in the first place.



\subsubsection{Grid Analysis on Stage I Information State}

\label{sec:grid}
While some prior work found that simple unimodal training failed to mitigate imbalance~\cite{wang2020makes}, we hypothesized that this was due to a lack of understanding of how the model's initial state affects joint training. To investigate this, we conducted a comprehensive grid analysis from an information-theoretic perspective to map out the initial information landscape across a wide range of Stage~II starting conditions. 

The detailed experiment settings are as follows, where we still use CREMA-D as the example. 
First, we trained separate unimodal classifiers for each modality (Audio and Visual) to create ``unimodal learning trajectories". We saved model checkpoints at various epochs along these trajectories, capturing the unimodal encoders' states as they updated. Considering that the audio modality learn much faster than the visual modality, we sampled checkpoints at non-uniform intervals, as shown in Figure~\ref{fig:pid_grid}. Then, we constructed a 5x5 grid of 25 unique starting points for joint training. Each point on this grid represents a specific combination of a pre-trained audio encoder and a pre-trained visual encoder (e.g., an audio encoder after 5 epochs combined with a visual encoder after 100 epochs). For each of these 25 initial states, we performed a FastPID analysis to measure the initial unique information (U1, U2), redundancy (R), and synergy (S).

The results, detailed in Figure~\ref{fig:pid_grid}, provide a comprehensive overview of how the initial information landscape is profoundly shaped by the unimodal pre-training duration.
\begin{itemize}
    \item \textbf{Diagnosing Inherent Imbalance with Uniqueness (U1, U2):} As shown in Figure~\ref{fig:pid_grid}, the grid clearly shows how different unimodal training durations create vast disparities in initial modality strength and how different paces the modalities learn. For instance, combining an audio encoder trained for 20 epochs with a visual encoder trained for 50 epochs results in an unexpected imbalance in unique information: U1 is 1.23 while U2 is a mere 0. This state is information-theoretically primed for the audio modality to dominate and suppress the visual one from the very start of joint training. This finding underscores the necessity of our controller's first function: to monitor the uniqueness ratio (U1/U2) and enforce balance by pausing the dominant modality, preventing such lopsided initial conditions.
    \item \textbf{Synergy (S) Reveals a Non-Monotonic ``Window of Opportunity":} Another critical insight comes from the behavior of synergy (S) in Figure~\ref{fig:pid_grid}, which quantifies the potential for multimodal fusion. Crucially, synergy is non-monotonic and does not simply increase with more unimodal training. Consider the row where the audio encoder is fixed at 10 epochs of pre-training. As the visual encoder's training progresses from 50 to 200 epochs, synergy first rises from 0.27 to a peak of 0.30 (at 100-150 visual epochs) before declining to 0.28. This pattern reveals an optimal "sweet spot" or "window of opportunity" for fusion. 
    Training a modality past this point in isolation can be detrimental to its collaborative potential, reducing its ability to synergize with the other modality. This empirical result provides the fundamental justification for our controller's function of tracking the peak of synergy (S) to adaptively and principledly determine the exact moment to transition from unimodal training to joint fusion.
    \item \textbf{Redundancy (R) increases along with training:} As expected, redundancy (R) generally increases as both modalities are trained for longer. This reflects that as both encoders become more proficient, they independently learn to capture the same obvious, shared information required for the classification task. 
\end{itemize}

In summary, \textit{this grid analysis moves beyond the simplistic conclusion that Stage I training is ineffective, but demonstrates that the initial information landscape is complex and highly tunable.} By using FastPID as a diagnostic tool, we can understand why certain initial states are superior. This provides the foundational evidence for our method, which navigates this landscape to find a starting point that is not just well-trained but information-theoretically balanced and primed for synergy.

\subsubsection{Correlation between Stage I PID metric and Fusion Performance} 
\label{sec:coeffient}
\begin{figure}[htp]
    \centering
    \includegraphics[width=0.99\linewidth]{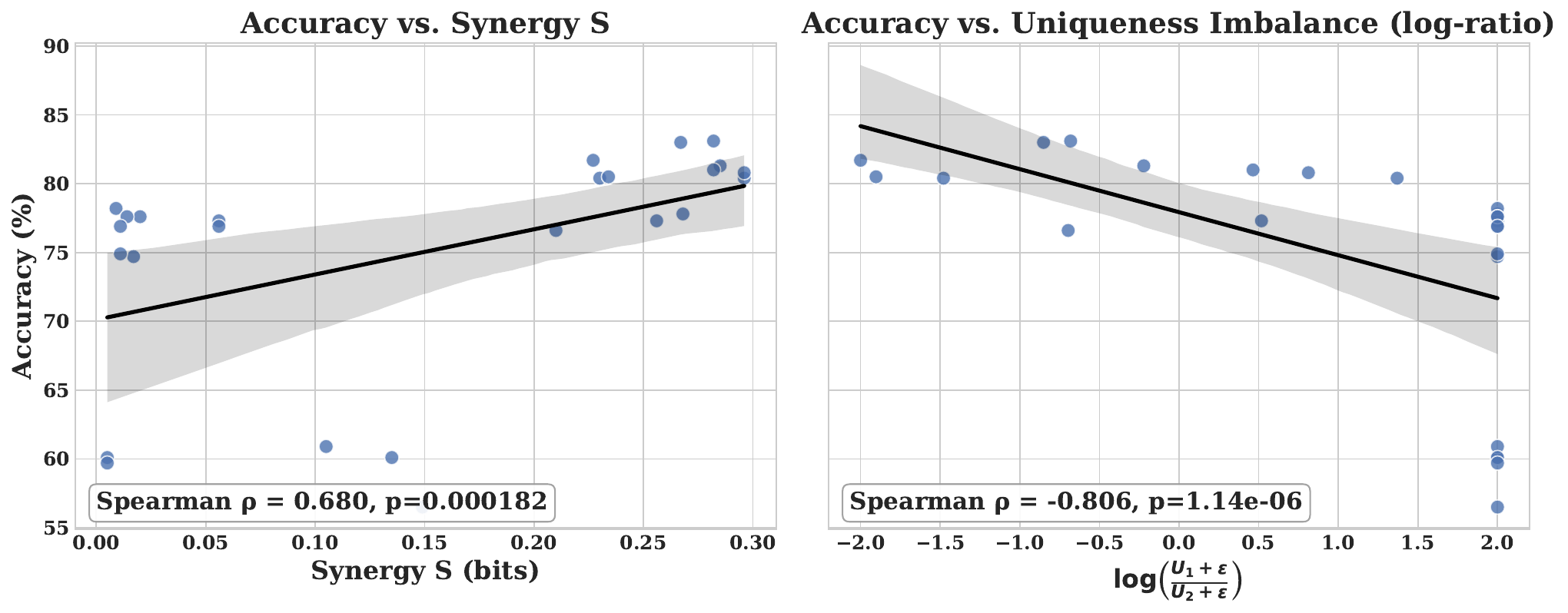}
    \caption{Validation of initial PID metrics as predictors of final task success. The scatter plots analyze the relationship between the initial information-theoretic state and the final classification accuracy on the CREMA-D dataset, across 25 unique starting points derived from our grid analysis.}
    \label{fig:spearman}
\end{figure}

To empirically validate our FastPID analysis, we measured the final task accuracy for each of the 25 initial states. The results (Table~\ref{tab:grid_accuracy}) confirm that the initial information-theoretic state strongly predicts final performance. The performance landscape exhibits a clear “sweet spot,” with a peak accuracy of \textbf{83.7\%} achieved at an asymmetrical starting point (Audio Epoch 5, Visual Epoch 200). This optimal point coincides with a region of high initial synergy identified by our PID analysis (Figure~\ref{fig:pid_grid}). A statistical analysis based on the scatter plots (Figure~\ref{fig:spearman}) further quantifies this link: initial synergy $S$ is strongly and positively correlated with final accuracy (Spearman $\rho=0.680$, $p=1.82\times10^{-4}$), whereas the log-ratio–based uniqueness imbalance is strongly and negatively correlated with accuracy (Spearman $\rho=-0.806$, $p=1.14\times10^{-6}$). These results substantiate our claim that initial states characterized by high synergy and low uniqueness imbalance are predictive of superior multimodal fusion..

Together, these results provide powerful empirical backing for our central claim: the initial PID metrics are not just theoretical constructs but strong predictors of final model performance.
This validates our FastPID-guided framework as a principled and efficient alternative to computationally expensive brute-force searches, enabling the proactive identification of the synergy-rich ``sweet spot'' that leads to synergetic multimodal fusion.

\begin{table}[htbp]
    \centering
    \caption{Final Prediction Accuracy (\%) for different Stage II initial states. The rows and columns correspond to the Stage I training epochs for the Audio and Visual encoders, respectively. The peak performance is highlighted in bold, and along with the underlined performances, formed a ``sweet spot".}
    \label{tab:grid_accuracy}
    \begin{tabular}{c|ccccc}
        \toprule
        \textbf{Audio Epoch} & \multicolumn{5}{c}{\textbf{Visual Epoch (\%)}} \\
        \cmidrule(l){2-6}
         & 1 & 50 & 100 & 150 & 200 \\
        \midrule
        1 & 56.5 & 76.6 & 80.4 & \underline{80.5} & \underline{81.7} \\
        5 & 60.1 & 77.3 & \underline{81.3} & \textbf{83.1} & \underline{83.0} \\
        10 & 60.9 & 77.8 & 80.4 & \underline{80.8} & \underline{81.0} \\
        20 & 60.1 & 74.7 & 77.3 & 78.2 & 77.6 \\
        25 & 59.7 & 74.9 & 76.9 & 77.6 & 76.9 \\
        \bottomrule
    \end{tabular}
\end{table}

\subsection{Comparison with MI-based Controller}
\label{sec:comparison_mi}

Our theoretical framework (Theorem~\ref{thm:mi_ecs_proxy}) establishes mutual information (MI), $I(Y; X_r)$, as a proxy for the unobservable Effective Competitive Strength (ECS). This raises a critical question: \textit{is the full complexity of Partial Information Decomposition (PID) necessary, or could a simpler controller guided only by MI suffice?} We argue that PID is essential because MI has two fundamental limitations: 1) it conflates unique and redundant information, providing a flawed signal for balancing modality competition, and 2) it is blind to cross-modal synergy, making it incapable of identifying the optimal moment for fusion. 

To empirically test this, we replaced our FastPID-guided asynchronous controller with a simpler version guided only by conventional MI. This MI-based controller uses the ratio $I(Y; X_1) / I(Y; X_2)$ to balance modalities and triggers the transition to Stage II when the sum $I(Y; X_1) + I(Y; X_2)$ peaks and begins to decline, using this as a proxy for optimal synergistic potential. 

The results, shown in Table~\ref{tab:pid_vs_mi_comparison}, confirm the inferiority of this approach. First, the MI ratio is an inaccurate measure of competitive strength because it fails to distinguish a modality's unique contribution from redundant, easily learned information. This led to a suboptimal training schedule, achieving an accuracy of only 79.70\%, significantly below the 83.50\% achieved by our controller. Second, the MI sum is a flawed proxy for synergy, as it conflates true collaboration with simple redundant learning. This caused the controller to transition prematurely, resulting in a final accuracy of just 72.28\%. These results demonstrate that being blind to synergy and unable to disentangle unique information leads to a poorly chosen initial state for Stage II, critically undermining model performance. Thus, the fine-grained analysis provided by FastPID is not a superfluous complexity but a necessary tool for effectively mitigating modality competition.

\begin{table}[htp]
\centering
\caption{Comparison of PID-guided vs. MI-guided control for determining the Stage II transition on CREMA-D.}
\label{tab:pid_vs_mi_comparison}
\begin{tabular}{l|c}
\toprule
\textbf{Method} & \textbf{Acc. (\%)} \\
\midrule
Unimodal Audio & 59.37 \\
Unimodal Visual & 30.46 \\
Concat Fusion & 62.68 \\
\hline
Controller with MI Ratio + Peak  & 72.28 \\
Controller with MI Ratio & 79.70 \\
\textbf{Controller with Uniqueness Ratio } & \textbf{82.20} \\
\textbf{Controller with Uniqueness Ratio + Synergy(Ours)} & \textbf{83.50} \\
\bottomrule
\end{tabular}
\end{table}

\subsection{Compatibility with Stage-II Balancing Methods}

We further evaluate the scalability and compatibility of our proposed framework with other existing methods that work on the modality competition. We hypothesize that by proactively creating an information-theoretically balanced and synergy-rich initial state, our framework, to a great extent, eases the modality competition problem before it emerges, rendering subsequent reactive interventions during joint training redundant or even slightly detrimental.

To validate this, we designed an experiment to evaluate the compatibility of our approach. We first apply our FastPID-guided asynchronous Stage I training to prepare the unimodal encoders. Then, instead of using a standard joint fusion, we apply established Stage~II balancing methods, specifically OGM-GE~\cite{peng2022balanced}, MMPareto~\cite{weiMMParetoBoostingMultimodal2024}, and ReconBoost~\cite{hua2024reconboost}, during the Stage~II fusion. We compare the performance of these hybrid models against our standalone framework and the original baselines on the CREMA-D dataset.

\begin{table}[htp]
\centering
\caption{Performance gain on CREMA-D dataset when applying our Stage I modulation on existing Stage II methods (\%). Our standalone method is shown to be superior, suggesting that reactive Stage II methods are somehow redundant when the initial state is properly conditioned.}
\label{tab:integration_crema_d}
\begin{tabular}{l|c}
\toprule
\textbf{Method} & \textbf{CREMA-D Acc. (\%)} \\
\midrule
Unimodal Audio & 59.37 \\
Unimodal Visual & 30.46 \\
\hline
Concat Fusion & 62.68 \\
OGM-GE~\cite{peng2022balanced} & 68.95 \\
MMPareto~\cite{weiMMParetoBoostingMultimodal2024} & 75.13 \\
ReconBoost~\cite{hua2024reconboost} & 79.82 \\
\hline
\textbf{Ours} & \textbf{84.27} \\
Ours + OGM-GE~\cite{peng2022balanced} & 83.74 \\
Ours + MMPareto~\cite{weiMMParetoBoostingMultimodal2024}  & 83.20 \\
Ours + ReconBoost~\cite{hua2024reconboost} & 83.20 \\
\bottomrule
\end{tabular}
\end{table}

The results, presented in Table~\ref{tab:integration_crema_d}, provide compelling evidence for our hypothesis. Our standalone framework, which combines our proactive Stage I with a simple joint fusion, achieves the highest performance at \textbf{84.27\%}, significantly outperforming all baselines.
Crucially, when our Stage I pre-training is combined with advanced Stage II methods, we observe no further performance gain. In fact, the performance of these hybrid models is slightly lower (83.20\% - 83.74\%) than our standalone method. This outcome strongly suggests that our proactive approach is not merely another component to be stacked, but a more fundamental solution.

\subsection{Evaluating FastPID: Accuracy and Speed}
\label{sec:pid_analysis}
While specialized conic solvers like SCS~\cite{o2016conic} offer high precision for PID, their significant computational cost and non-differentiable nature make them incompatible with end-to-end deep learning frameworks. To overcome these limitations, our FastPID combines a principled, closed-form analytical proxy for a highly informative warm-start with a fully differentiable refinement phase. This design sacrifices a negligible amount of numerical precision for dramatic gains in computational speed and full differentiability. This trade-off is what makes FastPID a practical, real-time diagnostic tool for guiding deep MML.

We rigorously validate FastPID’s efficacy through a multi-faceted computational analysis. We first benchmark its accuracy and scalability against a high-precision CVX solver~\cite{liang2024quantifying} on synthetic datasets (Section~\ref{sec:pid_syn}). We then demonstrate its practical utility and efficiency on the real-world CREMA-D dataset (Section~\ref{sec:pid_real}). Finally, we justify our design choices through ablation studies on both the analytical initialization (Section~\ref{sec:pid_ini}) and the refinement phase settings (Section~\ref{sec:pid_refine}).

\begin{table*}[htp]
\caption{Accuracy comparison of PID estimators on synthetic bitwise data where a theoretical ground truth (GT) exists. Our method (FastPID) is compared against the high-precision CVX solver~\cite{liang2024quantifying}.}
\label{tab:pid_bitwise_analysis}
\centering
\begin{tabular*}{\textwidth}{@{\extracolsep{\fill}}ll l cccc c@{}}
\toprule
\multirow{2}{*}{\textbf{Experiment}} & \multirow{2}{*}{\textbf{Task}} & \multirow{2}{*}{\textbf{Method}} & \multicolumn{4}{c}{\textbf{PID Values (bits)}} & \multirow{2}{*}{\textbf{MAE}} \\
\cmidrule(lr){4-7}
& & & \textbf{R} & \textbf{U\textsubscript{1}} & \textbf{U\textsubscript{2}} & \textbf{S} & \\
\midrule

\multirow{9}{*}{\textbf{Bitwise}} & \multirow{3}{*}{XOR} 
 & Theoretical GT     & 0.0000 & 0.0000 & 0.0000 & 1.0000 & --- \\
 & & CVX Solver       & 0.0000 & 0.0000 & 0.0000 & 1.0000 & 4.49e-08 \\
 & & \textbf{FastPID} & 0.0000 & 0.0000 & 0.0000 & 1.0000 & 1.01e-06 \\
\cmidrule(l){2-8} 

 & \multirow{3}{*}{AND} 
 & Theoretical GT     & 0.3100 & 0.0000 & 0.0000 & 0.5000 & --- \\
 & & CVX Solver       & 0.3111 & 0.0000 & 0.0002 & 0.4998 & 3.76e-04 \\
 & & \textbf{FastPID} & 0.3079 & 0.0033 & 0.0009 & 0.4991 & 1.79e-03 \\
\cmidrule(l){2-8}

 & \multirow{3}{*}{OR} 
 & Theoretical GT     & 0.3100 & 0.0000 & 0.0000 & 0.5000 & --- \\
 & & CVX Solver       & 0.3111 & 0.0005 & 0.0000 & 0.4997 & 4.75e-04 \\
 & & \textbf{FastPID} & 0.3082 & 0.0034 & 0.0009 & 0.4988 & 1.83e-03 \\
\bottomrule
\end{tabular*}
\end{table*}

\begin{table*}[htp]
\caption{Scalability and performance comparison of PID estimators on synthetic Gaussian data. As the number of input discretization bins ($D_x$) increases while output bins ($D_y$) are held constant, the problem size grows, testing the efficiency of each method. Error is reported as MAE against the CVX solver results (used as a reference).}
\label{tab:pid_gaussian_analysis}
\centering
\begin{tabular*}{\textwidth}{@{\extracolsep{\fill}}ll l cccc cc@{}}
\toprule
\multirow{2}{*}{\textbf{Experiment}} & \multirow{2}{*}{\textbf{Task}} & \multirow{2}{*}{\textbf{Method}} & \multicolumn{4}{c}{\textbf{PID Values (bits)}} & \multirow{2}{*}{\textbf{MAE}} & \multirow{2}{*}{\textbf{Time (s)}} \\
\cmidrule(lr){4-7}
& & & \textbf{R} & \textbf{U\textsubscript{1}} & \textbf{U\textsubscript{2}} & \textbf{S} & & \\
\midrule

\multirow{6}{*}{\textbf{Gaussian}} & \multirow{2}{*}{$D_x$=8, $D_y$=8} 
 & CVX & 0.1980 & 0.0850 & 0.0224 & 0.3265 & --- & 3.39 \\
 & & \textbf{FastPID} & 0.2081 & 0.0827 & 0.0215 & 0.3195 & 5.09e-03 & \textbf{0.32} \\
\cmidrule(l){2-9}

 & \multirow{2}{*}{$D_x$=16, $D_y$=8} 
 & CVX & 0.3097 & 0.0933 & 0.0340 & 0.6162 & --- & 10.60 \\
 & & \textbf{FastPID} & 0.3361 & 0.0883 & 0.0279 & 0.6010 & 1.32e-02 &\textbf{ 0.42} \\
\cmidrule(l){2-9}

 & \multirow{2}{*}{$D_x$=32, $D_y$=8} 
 & CVX & 0.3443 & 0.1221 & 0.0305 & 0.8464 & --- & 34.21 \\
 & & \textbf{FastPID} & 0.4169 & 0.1095 & 0.0105 & 0.8090 & 3.62e-02 & \textbf{0.55} \\
\bottomrule
\end{tabular*}
\end{table*}

\begin{figure}[htp]
    \centering
    \begin{subfigure}[b]{0.45\textwidth}
        \includegraphics[width=\textwidth]{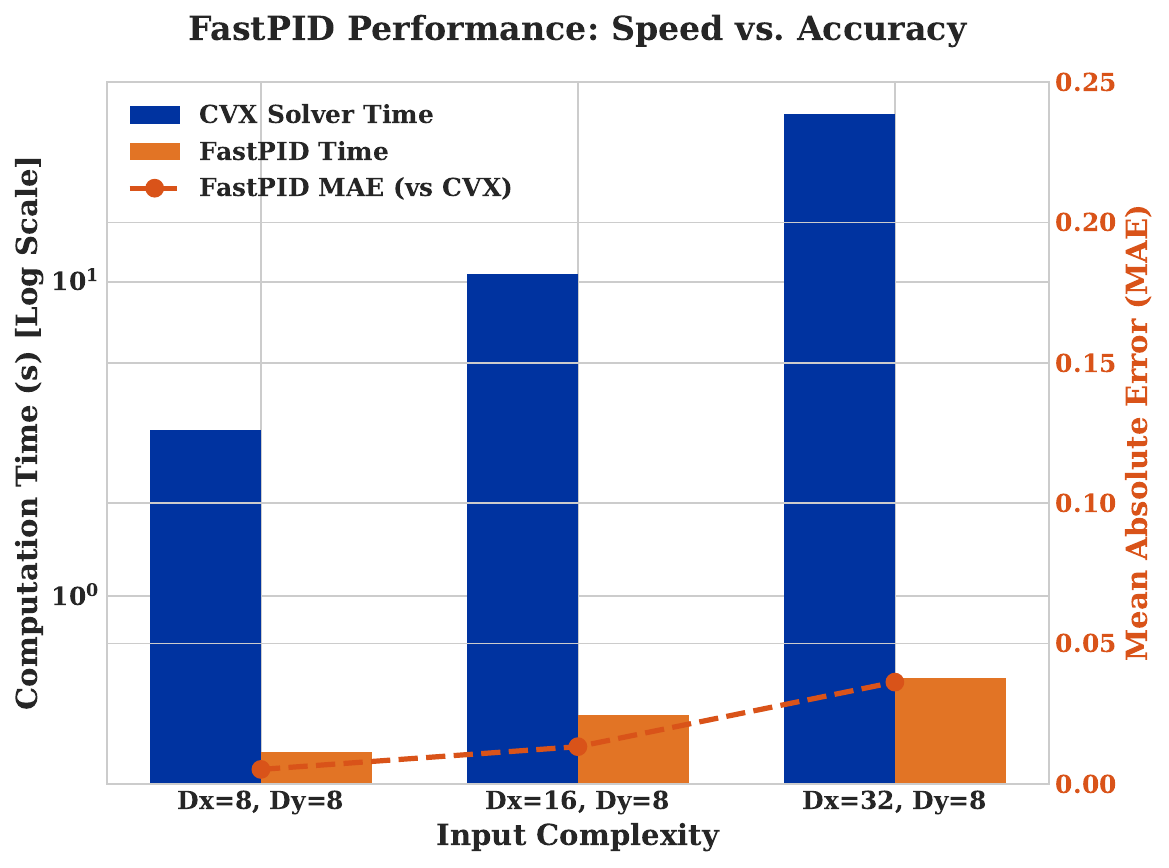}
        \caption{Performance vs. Input Complexity.}
        \label{fig:gaussian_performance}
    \end{subfigure}
    \hfill 
    \begin{subfigure}[b]{0.45\textwidth}
        \includegraphics[width=\textwidth]{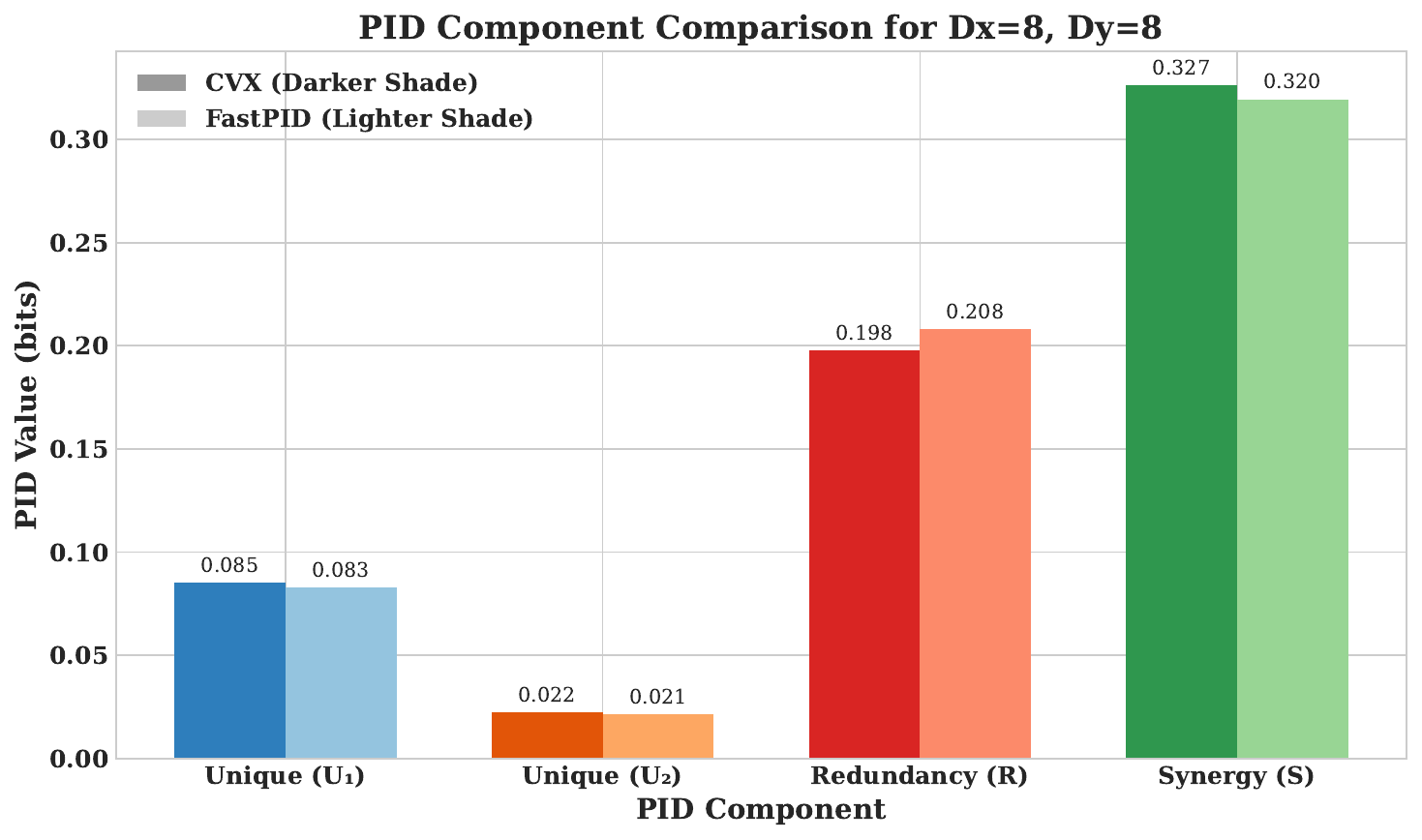}
        \caption{PID Component Breakdown (Dx=8, Dy=8).}
        \label{fig:gaussian_components}
    \end{subfigure}
    \caption{
        \textbf{Visual comparison of FastPID and the CVX solver on synthetic Gaussian data.} 
        \textbf{(a)} FastPID's computation time (orange bars) is orders of magnitude lower than CVX's (blue bars) and scales more efficiently with increasing input complexity. Simultaneously, its Mean Absolute Error (red line, right axis) remains minimal. 
        \textbf{(b)} A detailed comparison for the $D_x=8$ task shows that FastPID accurately recovers the individual PID components (R, U1, U2, S) with high fidelity compared to the CVX ground truth.
    }
    \label{fig:gaussian_analysis}
\end{figure}

\subsubsection{Accuracy and Scalability on Synthetic Data}
\label{sec:pid_syn}

To rigorously evaluate FastPID, we test it on two synthetic datasets designed to validate its two primary advantages: accuracy and computational efficiency. We assess its performance using two key metrics: (1) \textbf{Accuracy}, measured by the Mean Absolute Error (MAE) against a theoretical ground truth or a high-precision CVX solver as a reference, and (2) \textbf{Computation Time}, measured in seconds.

\noindent
\textbf{Synthetic Bitwise Data.} We first evaluate the methods on canonical bitwise operations (AND, OR, XOR), which are ideal benchmarks as their true PID values are theoretically known.Specifically, we generate discrete samples where $x_1, x_2 \sim \{0, 1\}$ and the label $y$ is determined by the respective logical gate ($y = x_1 \land x_2$, $y = x_1 \lor x_2$, $y = x_1 \oplus x_2$). 

As shown in Table~\ref{tab:pid_bitwise_analysis}, our FastPID solver demonstrates high accuracy. While the CVX solver achieves near-perfect precision as expected, FastPID produces results with a mean absolute error that is negligibly small (on the order of $10^{-3}$). Also, it correctly identifies the pure synergy (1.0 bits) in the XOR task and the dominant redundancy ($\approx$0.31 bits) in the AND/OR tasks.

\noindent
\textbf{Synthetic Gaussian Data.} To simulate the continuous and noisy feature embeddings from real-world encoders, we designed an experiment using synthetic Gaussian data target variable $Y$ from a mix of linear ($c_1X_1 + c_2X_2$) and non-linear quadratic ($u_1X_1^2 + u_2X_2^2$) components, with added Gaussian noise. To test scalability, we varied the number of discretization bins for the inputs, $D_x \in \{8, 16, 32\}$, while keeping the target's bins fixed ($D_y=8$).

The results, summarized in Table~\ref{tab:pid_gaussian_analysis} and Figure~\ref{fig:gaussian_analysis}, highlight FastPID's primary advantages: exceptional computational efficiency and high accuracy. As input complexity increases, FastPID's runtime scales far more gracefully than the CVX solver, achieving a \textbf{speedup of 10x to over 60x}. Crucially, this efficiency does not compromise accuracy: the MAE relative to the CVX reference remains negligibly small. A detailed component-wise comparison confirms that FastPID accurately recovers the individual PID values (Redundancy, Uniqueness, and Synergy). These results prove that FastPID is a highly scalable and accurate tool, making it practical for analyzing the high-dimensional feature spaces encountered in deep learning.

\begin{figure*}[t]
    \centering
    \includegraphics[width=0.95\textwidth]{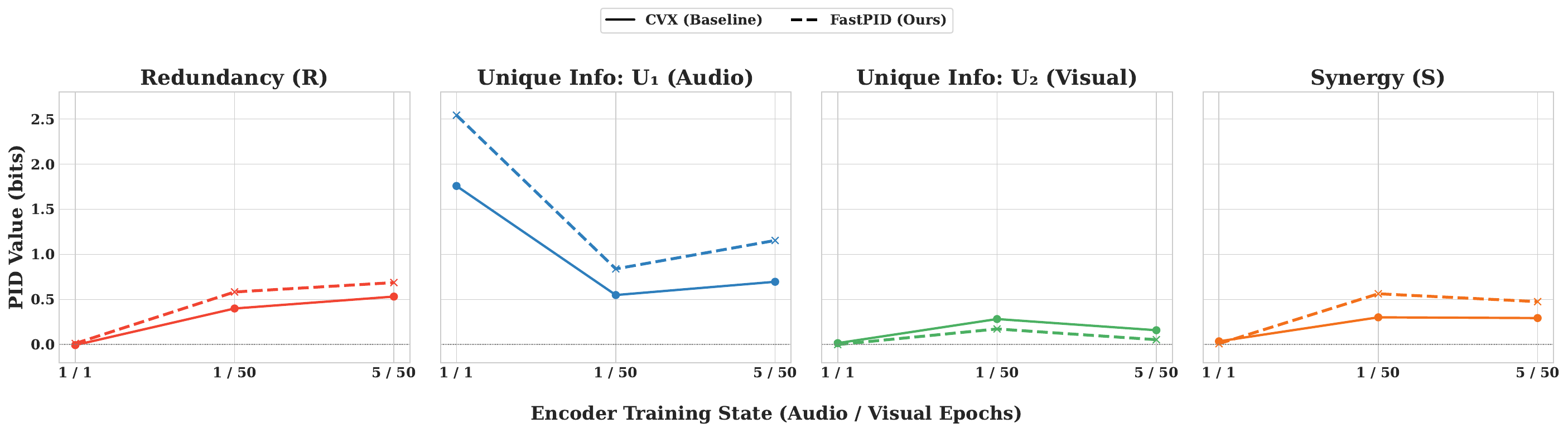}
    \caption{
        This figure complements Table~\ref{tab:real_data_trend_comparison} by visualizing the information-theoretic trends. Each panel corresponds to a single PID atom (R, $U_1$, $U_2$, S). The solid lines represent the CVX baseline, while the dashed lines represent our FastPID method. The close tracking between the two lines across all three training scenarios visually confirms that FastPID reliably captures the same complex dynamics, from audio dominance to collaboration and back to competition, as the much slower baseline method.
    }
    \label{fig:real_data_trends}
\end{figure*}

\subsubsection{Applicability and Efficiency on Real-World Data}
\label{sec:pid_real}
To demonstrate FastPID's practical applicability and efficiency, we extend our analysis to the real-world CREMA-D dataset, aiming to understand the interplay between audio ($X_1$) and visual ($X_2$) modalities. We quantize the high-dimensional feature embeddings from each unimodal encoder into $k=20$ discrete categories, constructing a $20 \times 20 \times 6$ joint probability distribution $P(X_{audio}, X_{visual}, Y)$ to be processed.

Similar to the grid analysis in Section~\ref{sec:grid}, we probe the model at different unimodal training epochs, computing PID values using both our proposed FastPID and a baseline CVX solver. The comparison focuses on two primary aspects: (1) \textbf{Computation Time}, to demonstrate efficiency, and (2) \textbf{Result Consistency}, to show that our method reliably tracks the same information-theoretic trends as the baseline. Note that given the instability of the CVXPY solver on such large-scale problems, we use its results as a reference baseline rather than an absolute ground truth.

\noindent
\textbf{Results and Analysis.}
Our experiments on real-world data demonstrate that FastPID \textit{transforms the computationally prohibitive task of tracking information dynamics into a practical diagnostic tool for MML models.} Its core advantage is its ability to efficiently uncover the evolution of a model's information-processing strategy, a task intractable with traditional methods.

The results, summarized in Table~\ref{tab:real_data_trend_comparison} and Figure~\ref{fig:real_data_trends}, reveal a compelling narrative of modality collaboration and competition:
\begin{itemize}
    \item \textbf{Nascent State (Epoch 1/1):} Initially, the system is severely imbalanced, with information being almost entirely unique to the audio modality ($U_1$) and negligible synergy ($S$).
    
    \item \textbf{Asymmetric Training (Epoch 1/50):} As the visual model matures, a desirable shift to \textbf{multimodal collaboration} occurs, marked by a dramatic increase in both Redundancy and Synergy.
    
    \item \textbf{Modality Competition (Epoch 5/50):} The collaborative state is fragile: A small amount of additional audio training rapidly triggers modality competition, where \textbf{the progress of 50 visual epochs is reversed by just 5 audio epochs}. This is seen in a sharp rebound in audio uniqueness ($U_1$) and a collapse in synergy ($S$), indicating the model is unlearning its effective fusion strategy.
\end{itemize}

This entire diagnostic narrative is made possible by FastPID's efficiency. Being \textbf{100x to 1000x faster} than the CVX baseline, it allows for fine-grained temporal analysis, turning PID from a theoretical concept into a practical debugging tool for complex multimodal systems.

\begin{table*}[t]
\centering
\caption{
    \textbf{Tracking information dynamics during multimodal model training.} FastPID consistently captures the same trends in information dynamics as the CVX baseline, but at a fraction of the computational cost (\textbf{100x-1000x speedup}). As the models progress, both methods detect significant shifts from audio dominance to multimodal collaboration and finally to modality competition, demonstrating FastPID's reliability for practical model analysis.
}
\label{tab:real_data_trend_comparison}
\resizebox{\textwidth}{!}{%
\begin{tabular}{@{}ll cccccc@{}}
\toprule
\textbf{Training Scenario} & \textbf{Epochs (A/V)} & \textbf{Method} & \textbf{R} (Redundancy) & \textbf{U\textsubscript{1}} (Audio) & \textbf{U\textsubscript{2}} (Visual) & \textbf{S} (Synergy) & \textbf{Time (s)} \\
\midrule

\multirow{2}{*}{\begin{tabular}[c]{@{}l@{}}\textbf{Nascent State:} \\ Untrained Models\end{tabular}} & \multirow{2}{*}{1 / 1} 
& CVX (Baseline) & -0.006 & 1.758 & 0.016 & 0.036 & 866.4 \\
& & \textbf{FastPID (Ours)} & \textbf{0.012} & \textbf{2.542} & \textbf{0.000} & \textbf{0.010} & \textbf{0.8} \\
\cmidrule(l){2-8}

\multirow{2}{*}{\begin{tabular}[c]{@{}l@{}}\textbf{Asymmetric Training:} \\ Visual Model Matures\end{tabular}} & \multirow{2}{*}{1 / 50} 
& CVX (Baseline) & \underline{0.400} & 0.549 & 0.283 & \underline{0.302} & 855.9 \\
& & \textbf{FastPID (Ours)} & \underline{\textbf{0.583}} & \textbf{0.839} & \textbf{0.173} & \underline{\textbf{0.563}} & \textbf{7.5} \\
\cmidrule(l){2-8}

\multirow{2}{*}{\begin{tabular}[c]{@{}l@{}}\textbf{Modality Competition:} \\ Audio Dominance Emerges\end{tabular}} & \multirow{2}{*}{5 / 50} 
& CVX (Baseline) & 0.532 & 0.696 & 0.159 & 0.294 & 851.5 \\
& & \textbf{FastPID (Ours)} & \textbf{0.687} & \textbf{1.154} & \textbf{0.054} & \textbf{0.476} & \textbf{10.1} \\
\bottomrule
\end{tabular}
}
\end{table*}

\begin{table}[h!]
\centering
\caption{Ablation study on initialization methods where we compare the proposed analytical start against uniform and Gaussian initializations.}
\label{tab:ablation_initialization}
\begin{tabular}{l|l|r|r|r}
\hline
\textbf{Initialization} & \textbf{Solver} & \textbf{Time (s)} & \textbf{MAE} & \textbf{Iters/Status} \\ \hline
\textbf{Analytical} & \textbf{CVX} & \textbf{17.22} & \textbf{0.00} & \textbf{N/A} \\
\textbf{Analytical} & \textbf{FastPID} & \textbf{3.42} & \textbf{3.17e-02} & \textbf{419} \\ \hline
Uniform & CVX & 17.26 & 0.00 & N/A \\
Uniform & FastPID & 16.58 & 5.55e-02 & 2000\\ \hline
Gaussian & CVX & 16.77 & 0.00 & N/A \\
Gaussian & FastPID & 9.01 & 6.42e-02 & 959 \\ \hline
\end{tabular}
\end{table}

\subsubsection{FastPID Analysis on Initialization Points}
\label{sec:pid_ini}
A core contribution of our FastPID solver is its hybrid design, which begins with a closed-form analytical proxy for a \textit{warm-start}. To validate this choice, we conducted an ablation study comparing our \textbf{Analytical} initialization against \textbf{Uniform} and \textbf{Gaussian} starting points on the synthetic Gaussian dataset.

We test these initializations on both our FastPID solver and the baseline CVXPY solver using the synthetic Gaussian dataset described in Section~\ref{sec:pid_syn}. The results, presented in Table~\ref{tab:ablation_initialization}, offer a clear justification for our approach. While the high-precision CVX solver's performance is unaffected by the starting point, remaining consistently slow (~17s), our FastPID solver's performance is highly sensitive to it. When warm-started with the analytical proxy, FastPID is both fast and accurate, converging in just 3.42 seconds (a \textbf{5x speedup} over CVX) with the lowest Mean Absolute Error (MAE). In contrast, starting from a uniform or Gaussian distribution dramatically degrades performance, leading to significantly longer computation times and higher error, with the uniform start failing to converge within the iteration limit.

This analysis confirms that the analytical initialization is a critical component of FastPID's efficiency. By providing a starting point that is already close to the optimal solution, it enables the gradient-based refinement to converge rapidly and accurately, making FastPID a practical and scalable tool for deep learning applications.

\subsubsection{Ablation Study on Different Refinement Settings}
\label{sec:pid_refine}

To validate the design of FastPID's refinement phase, we conducted an ablation study on its key hyperparameters using the synthetic Gaussian dataset similar in Section~\ref{sec:pid_ini}, with default settings detailed in Table~\ref{tab:ablation_defaults}.

\begin{table}[h!]
\centering
\caption{Default settings for the synthetic data generation and the FastPID solver used throughout the ablation study.}
\label{tab:ablation_defaults}
\begin{tabular}{ll}
\hline
\textbf{Parameter} & \textbf{Default Value} \\ \hline
\textit{-- Data Generation --} & \\
Number of Samples & 500,000 \\
Input Bins (\texttt{n\_bins\_x}) & 16 \\
Target Bins (\texttt{num\_y\_bins}) & 8 \\
\hline
\textit{-- Solver Settings --} & \\
Max Iterations (\texttt{max\_iter}) & 2000 \\
Learning Rate (\texttt{lr}) & 0.1 \\
Tolerance (\texttt{tol}) & 1e-5 \\
Sinkhorn Iterations (\texttt{max\_sinkhorn\_iter}) & 10 \\ \hline
\end{tabular}
\end{table}

\noindent
\textbf{Refinement Iterations.}
We first evaluated the effect of the maximum number of refinement steps (\texttt{max\_iter}). The results are shown in Table~\ref{tab:ablation_max_iter}. Note that the MAE does not monotonically decrease with more iterations: the lowest error is achieved with just 50 iterations, yielding a 42.3x speedup. This behavior is characteristic of gradient-based optimization where the fixed learning rate causes the optimizer to overshoot the minimum and diverge, increasing the MAE. This finding is significant: it shows that FastPID requires only a brief refinement phase to achieve high accuracy, highlighting the effectiveness of its principled warm-start.

\begin{table}[h!]
\centering
\caption{Ablation results for the number of iterations.}
\label{tab:ablation_max_iter}
\begin{tabular}{cccc}
\hline
\textbf{Max Iters} & \textbf{Time (s)} & \textbf{Speedup} & \textbf{MAE (vs CVX)} \\ \hline
10 & 0.1000 & 162.4x & 3.76e-02 \\
50 & 0.3844 & 42.3x & 7.50e-03 \\
200 & 1.4895 & 10.9x & 2.05e-02 \\
500 & 3.9411 & 4.1x & 3.63e-02 \\
2000 & 19.6247 & 0.8x & 7.10e-02 \\ \hline
\end{tabular}
\end{table}

\noindent
\textbf{Sinkhorn Iteration Step.}
Next, we analyzed the impact of the number of Sinkhorn projection iterations (\texttt{max\_sinkhorn\_iter}). The results in Table~\ref{tab:ablation_sinkhorn} reveal a critical interplay: investing more computation in the projection step can drastically reduce the total number of outer loop iterations required. This is because with too few Sinkhorn iterations, the projection is inaccurate, forcing the main solver to perform many more steps to converge. The best performance (lowest MAE and a 6.7x speedup) was achieved with 20 Sinkhorn iterations, which allowed the solver to converge in the fewest actual iterations by making each projection more effective.

\begin{table}[h!]
\centering
\caption{Ablation results for the number of Sinkhorn iterations.}
\label{tab:ablation_sinkhorn}
\begin{tabular}{ccccc}
\hline
\textbf{Sinkhorn Iters} & \textbf{Time (s)} & \textbf{Speedup} & \textbf{MAE} & \textbf{Actual Iters} \\ \hline
1 & 1.61 & 10.1x & 5.77e-02 & 236 \\
5 & 2.93 & 5.6x & 7.364e-02 & 600 \\
10 & 9.82 & 1.7x & 5.88e-02 & 940 \\
20 & 2.42& 6.7x & 7.40e-03 & 117 \\
40 & 4.69 & 3.5x & 7.40e-03 & 116 \\ \hline
\end{tabular}
\end{table}

\section{Discussion and Conclusion}

Different from the prevailing approach that deals with modality competition during the single-stage joint training process, in this work, we propose a two-stage framework that adopts the unimodal training scheduling to shape the model's initial state for synergistic fusion. Our primary theoretical contribution introduces Effective Competitive Strength (ECS), revealing a "Competition Breaking" state that could achieve a provably tighter test error bound. To operationalize this theory, we developed a FastPID-guided asynchronous controller that balances modalities by monitoring uniqueness and identifies the ideal fusion point by tracking peak synergy, enabled by FastPID, a computationally efficient and differentiable Partial Information Decomposition solver that transforms PID into a real-time diagnostic tool. Comprehensive experiments validate our approach, achieving state-of-the-art results with an average gain of 7.70\% across four diverse benchmarks.

Future work will focus on extending our framework to scenarios with more than two modalities, adapting it for self-supervised learning, and leveraging the differentiability of FastPID as a direct training regularizer. Ultimately, this work advocates for a paradigm shift in multi-modal learning: from reactively managing competition to proactively engineering initial states for collaboration, thereby laying a more robust foundation for integrated AI systems.


\bibliographystyle{IEEEtran}
\bibliography{cite}


\begin{IEEEbiographynophoto}{Jiaqi Tang} received the bachelor’s degree from the Beijing University of Posts and Telecommunications. 
Currently, he is working towards a Ph.D. degree at Peking University. He is also affiliated with the National Institute of Health Data Science, Peking University.
\end{IEEEbiographynophoto}


\begin{IEEEbiographynophoto}{Yinsong Xu} received the bachelor’s degree from the Beijing University of Posts and Telecommunications. He
is working toward the PhD degree at the Beijing University of Posts and Telecommunications. He is also affiliated with the National Institute of Health Data Science, Peking University.
\end{IEEEbiographynophoto}


\begin{IEEEbiographynophoto}{Yang Liu} received PhD and MPhil in Advanced Computer Science from University of Cambridge, and B.Eng. in Telecommunication Engineering from Beijing University of Posts and Telecommunications (BUPT). She is now a Tenure-track Assistant Professor (Ph.D. Supervisor) in Wangxuan Institute of Computer Technology, Peking University/
\end{IEEEbiographynophoto}


\begin{IEEEbiographynophoto}
{Qingchao Chen} received the B.Sc. degree in telecommunication engineering from the Beijing University of Post and Telecommunication and the Ph.D. degree from the University College London. He was a Postdoctoral Researcher with University of Oxford, U.K. (2018-2021). He is currently an assistant professor at the National Institute of Health Data Science, Peking University. His current researches focus on computer vision and machine learning, radio-frequency signal
processing and system design, and biomedical multimodality data analysis.
\end{IEEEbiographynophoto}

\end{document}